%% file: main.tex
\documentclass[conference]{IEEEtran}

\usepackage[normalem]{ulem}
\usepackage{amsmath}
\usepackage{amsfonts}

\usepackage{xcolor}
\usepackage[pdftex]{graphicx}
\graphicspath{{images/}}
\usepackage{float}
\usepackage{caption, subcaption}
\usepackage[export]{adjustbox}
\usepackage[leftcaption]{sidecap} 
\usepackage{wrapfig}

\usepackage{amssymb}% http://ctan.org/pkg/amssymb
\usepackage{pifont}% http://ctan.org/pkg/pifont
\newcommand{\cmark}{\ding{51}}%
\newcommand{\xmark}{\ding{55}}%
\usepackage{multirow}
\usepackage{multicol}
\usepackage{arydshln}
\usepackage[edges]{forest}
\usepackage{nccmath}

\newcommand{\eat}[1]{}

\renewcommand{\arraystretch}{1.5}

\usepackage[official]{eurosym}
\usepackage{enumitem}
\setlist{nolistsep,noitemsep}
\usepackage{hyperref}

\begin{document}

%\title{Survey of auto-features engineering for times series classification -  evaluation and discussion}

\title{Automatic Feature Engineering\\for Time Series Classification:\\ Evaluation and Discussion}


 \author{\IEEEauthorblockN{Aurélien Renault}
 \IEEEauthorblockA{Orange Innovation Paris, France\\
 Email: aurelien.renault@orange.com} \\
 \IEEEauthorblockN{Vincent Lemaire}
 \IEEEauthorblockA{Orange Innovation, Lannion, France\\ 
 Email: vincent.lemaire@orange.com}
 \and
 \IEEEauthorblockN{Alexis Bondu}
 \IEEEauthorblockA{Orange Innovation, Paris, France\\
 Email: alexis.bondu@orange.com}\\ 
 \IEEEauthorblockN{Dominique Gay}
 \IEEEauthorblockA{Université de la Réunion, France\\
 Email: dominique.gay@univ-reunion.fr}}
%%%%%%%%%%%%%%%%%%%%%

\maketitle

\begin{abstract}
Time Series Classification (TSC) has received much attention in the past two decades and is still a crucial and challenging problem in data science and knowledge engineering. Indeed, along with the increasing availability of time series data, many TSC algorithms have been suggested by the research community in the literature. Besides state-of-the-art methods based on similarity measures, intervals, shapelets, dictionaries, deep learning methods or hybrid ensemble methods, several tools for extracting unsupervised informative summary statistics, aka \emph{features}, from time series have been designed in the recent years. Originally designed for descriptive analysis and visualization of time series with informative and interpretable features, very few of these feature engineering tools have been benchmarked for TSC problems and compared with state-of-the-art TSC algorithms in terms of predictive performance. In this article, we aim at filling this gap and propose a simple TSC process to evaluate the potential predictive performance of the feature sets obtained with existing feature engineering tools. Thus, we present an empirical study of 11 feature engineering tools branched with 9 supervised classifiers over 112 time series data sets. The analysis of the results of more than 10000 learning experiments indicate that feature-based methods perform as accurately as current state-of-the-art TSC algorithms, and thus should rightfully be considered further in the TSC literature.% (+ phrase reproducible github).
%The Time series classification (TSC) literature is quite singular as most accurate algorithms exhibit fundamentally different approaches with very similar predictive performances. After presenting the TSC landscape, we propose a novel features-based approach. Indeed, many off-the-shelf tools can easily allow one user to extract statistics, transformations specifically designed for time series. Those are, so far, used for analysis and/or visualizations rather than for classification purpose. Considered both scalability and interpretability of those methods, we used those different libraries alongside some usual, non time series dedicated, classifier to see how well could performances be using those features. Combining several features extraction libraries, without hardly tuning any of the pipeline step, showed to perform as accurately as the current state-of-the-art, time series dedicated, classifier, with a significant speed-up in run time. These experiments led us think that features-based methods should rightfully be considered further in TSC literature.  .
\end{abstract}

%%%%%%%%%%%%%%%%%%%%%%%%%
\input{intro_v2}
\input{related_v2}
\input{experiments_v2}
\input{conclusion_v2}

\bibliographystyle{IEEEtran}
\bibliography{main.bbl}

\end{document}

%% file: intro_v2.tex
%auto-ignore
%===========================================================
%==================== INTRODUCTION =========================
%===========================================================

\section{Introduction}
The goal of Time Series Classification (TSC) is to assign a class label $y$ from a set $Y=\{y_i\}$ of predefined labels to an unlabeled time series $X = [x_1, x_2, \ldots, x_m]$ which is an ordered set of real values. The associated machine learning task is to train a classifier function $f$ on a labeled time series data set $\mathcal{D} = \{(X_1,y_1), (X_2,y_2), \ldots, (X_n,y_n)\}$ in order to map the space of possible input series to the class labels of $Y$. Then, for an incoming unlabeled time series $X$, the prediction of the assignment is given by $f(X)$.

Formulated as such, TSC has been identified as a top-10 challenging problem in data mining research~\cite{YW06} and has received much attention in the literature. This particular attention is due to the overwhelming amount of available time series data~\cite{EA12}. Indeed, in many scientific fields, measurements are taken over time. The resulting collection of ordered data is represented as time series. Thus, times series data arise from many real-world applications: e.g., in the UCR/UEA archive~\cite{BLVK}, audio signal, electrocardiogram, encephalogram, human activity, image or motion classification are among the most frequent tasks. In order to solve these tasks, hundreds of TSC algorithms have been suggested in the past two decades~\cite{bagnall2017great}. Various paradigms have been exploited in the TSC literature; from simple distance-based Nearest Neighbors to deep learning architectures and complex hybrid ensemble methods. 
%(we report a structured overview in the next section).
The very latest contributions~\cite{middlehurst2021hive} indicate that five methods, HIVECOTE (ensemble methods HC1~\cite{lines2018time} and HC2~\cite{middlehurst2021hive}), convolutional kernels based ROCKET~\cite{dempster2020rocket}, tree ensemble TS-CHIEF~\cite{shifaz2020ts} and deep learning based InceptionTime~\cite{ismail2020inceptiontime} achieve top predictive performance.

Besides the numerous TSC algorithms, another research area has been developed in the recent years, namely time series Feature Construction (FC). FC approaches are attractive as they regroup methods for extracting informative and interpretable summary statistics from time series. Features might be diverse to capture various properties of the time series, e.g., seasonality, trends, autocorrelation, etc, and thus can be adapted to various application domains. The pioneering work HCTSA~\cite{FLJ13} allows to extract more than 7000 unsupervised time series features summarizing properties of the distribution of values in a time series, correlation properties, entropy and complexity measures, how properties of a time series change over time, etc. Since then, several unsupervised feature engineering tools have been independently developed in various programming languages: e.g., C/Python-based CATCH22~\cite{lubba2019catch22}; Python-based FEATURETOOLS~\cite{kanter2015deep}, TSFRESH~\cite{christ2016distributed,CBN+18}, TSFEL~\cite{barandas2020tsfel}; and R-based  TSFEATURES~\cite{hyndman2019tsfeatures}, FEASTS~\cite{o2019package}. Originally designed for descriptive analysis and visualization of time series, their use for TSC problems has been less studied: e.g., B.~Fulcher et al.~\cite{fulcher2017hctsa} suggest a linear classifier based on the feature set generated by HCTSA and on a much smaller set, CATCH22~\cite{lubba2019catch22}. However, to our knowledge, no extensive study has been led on the effectiveness of existing feature engineering tools for TSC problems and how they compare with the leading TSC performers.

%\textcolor{blue}{AB - peut-être qu'on ne comprend pas assez bien comment se positionne l'apporche simple que nous proposons par rapport aux approches spécifiques de l'état de lart (VL pas sûr qu'il soit utile d'ajouter ce paragraphe en bleu): Most TSC approaches in the literature are composed of a feature extraction step and a classification step. In general, in these works, the relative importance of these two modeling steps in obtaining good performances is poorly studied. In this paper, we address this issue by assembling unsupervised feature extraction tools with standard classifiers for tabular data. We show that this simple approach is competitive with state of the art approaches, which are allowed to use labels during feature extraction and adapt the classification step to the particular type of data (i.e. time series).}

\begin{figure}[htbp!]
\centering
\includegraphics[width=1.0\columnwidth]{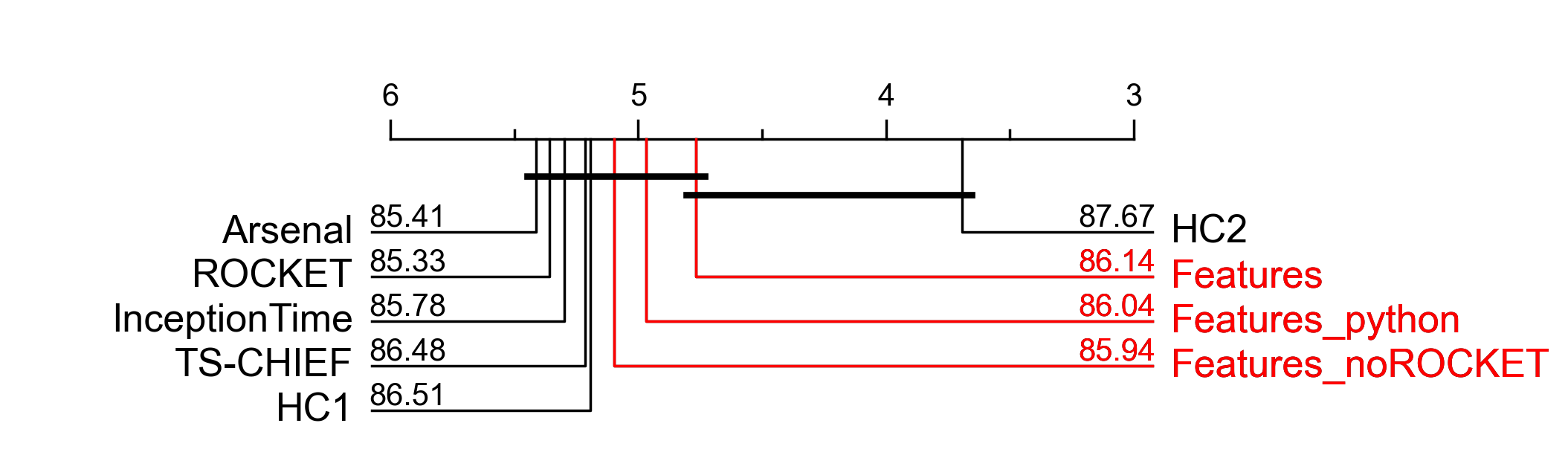}
\caption{Critical difference diagram of FC variants against current state of the art on 112 UCR/UEA data sets. The average rank is plotted and each contender is annotated by its mean accuracy results. Thick lines linking group of classifiers indicate no significant difference in predictive performance. This demonstrates that suggested FC processes (in red) are among the best TSC performers. }
\label{fig:sota-intro}
\end{figure}

In this article, we address these shortfalls and suggest several simple learning processes to assess the effectiveness of feature construction for TSC. Our two-step learning processes consist in : \textit{(i)} transforming the original time series data into feature-based tabular data using an unsupervised feature engineering tool (or a combination of tools), and \textit{(ii)} learning a supervised classifier on the obtained tabular data. Figure~\ref{fig:sota-intro} summarizes the final results of three selected processes compared with current leaders on 112 UEA/UCR data sets. Using feature construction is then comparable with the current state of the art TSC algorithms in terms of predictive performance (exhaustive details of the experiments are provided in Section~\ref{protocol}). %\textcolor{magenta}{+ phrase sur le compromis running time / accuracy ?}

The rest of the paper is organized as follows: in Section~\ref{literature review}, we position our feature construction approach w.r.t. related work in TSC and feature engineering. The empirical evaluation and comparison are reported in Section~\ref{protocol}. We conclude in Section~\ref{conclusion} with some perspectives for feature construction in TSC problems.

%% file: related_v2.tex
\section{Background} \label{literature review}
One of the simplest, but also very popular method for TSC problems is $k$-Nearest-Neighbors coupled with a similarity measure, e.g., Euclidean distance or dynamic time warping. Following these baselines, the TSC literature has grown tremendously. With the main focus on efficiency and predictive performance, several paradigms have been exploited:
\begin{itemize}
\item \textit{similarity-based:} generally coupled with 1-Nearest Neighbor (1-NN), several similarity measures have been experimented for TSC problems. Elastic similarity measures provide the best predictive performance as they allow to tackle with local distortions and misalignments in time series~\cite{LB15}.
\item \textit{interval-based:} from random intervals extracted from time series, a new feature vector is build and a traditional classifier is trained. Successful representatives of this type of method are the ones using multiple intervals and ensemble classifier, see e.g.~\cite{BR16}. 
\item \textit{shapelet-based:} the search for sub-sequences highlighting class separation not depending on when they occur in a serie has attracted many research efforts since~\cite{YK09}.
\item \textit{dictionary-based:} these algorithms build words from sliding windows, then a classifier is learned on the transformed data made of histograms of word counts~\cite{LBMT19}
\item \textit{deep learning based:} following the success of deep learning for images and videos, several neural network architectures have been suggested for TSC~\cite{ismail2019deep}.
\item \textit{ensemble-based:} to combine the performance of existing approaches, meta-ensemble classifiers have been designed and achieved top predictive performance. For example, the Collective of Transformation-based Ensembles (CoTE~\cite{BLHB15}) and its successors HC1~\cite{lines2018time} and HC2~\cite{middlehurst2021hive} fall in this category and are also providing structured overviews of the TSC literature.
\end{itemize}
As the TSC literature already offers well-structured overviews of existing TSC approaches, in the rest of the background part, we will concentrate on the recent top performers. The following algorithms are also the contenders in the experimental section. 
% \textcolor{magenta}{TO DO :\\
% -- UNE COLONNE MAX-- quick summary of the various paradigms in TSC, distance, intervals, shapelets, dictionaries, ensemble \ldots \\
% -- puis top performers auxquels on se compare, ci dessous à réduire}

\subsection{Top TSC performers}
\noindent \textbf{ROCKET family} One of the most promising method from the past few years is the ROCKET (Random Convolutional KErnel Transformation) model (\cite{dempster2020rocket}), which achieved SOTA performances with just a fraction of the time needed for competing approaches. The idea of ROCKET is to use a high number of random convolutional kernels with random lengths, dilations, padding and weights, then convolve them with the time series samples to finally extract two features per kernel, which are the max and the \textit{ppv} (proportion of positive values) for each sample. A linear classifier is then trained on generated features, either a Ridge or a Logistic regression depending on the data set size we are dealing with: logistic regression will generally be preferred working with bigger data sets.
%The parameters used to generate the kernels have been tuned on the considered benchmark, and thus can take values in a restricted set of possibilities, empirically limiting kernels complexity in order to avoid overfitting. Thus, this set of possible parameters' values may overfit on the benchmark, which could represent one limitation.

An extension of the ROCKET framework has been further proposed by the authors in MiniROCKET (\cite{dempster2021minirocket}), which is now used as the default ROCKET variant in most analysis. The goal of MiniROCKET is to reduce, and almost eradicate randomness from the pipeline. In order to do this, authors have restricted much more the kernels parameters and only extract the \textit{ppv} for each sample, making the algorithm almost deterministic. Hence, a pre-defined set of 84 kernels has been chosen (among the $512$ possible ones), which is supposed to balance accuracy and computational cost.
%Table \ref{fig:table_change} is displaying all the changes made from ROCKET.
Within this framework, MiniROCKET achieved about the same level of accuracy as ROCKET but way faster (up to 75x faster on larger data sets), making this approach even more competitive when it comes to computation time and scalability. 

% \begin{figure}[!htb]
%     \centering
%     \includegraphics[width=\linewidth]{figures/mini_changes.png}
%     \caption{Summary of changes of MiniROCKET \textcolor{magenta}{(from \cite{dempster2021minirocket})}}
%     \label{fig:table_change}
% \end{figure}

One more ROCKET variant has been developed, this time focusing on predictive performance rather than speed, namely MultiROCKET (\cite{tan2021multirocket}). This algorithm is significantly improving the overall performance by extracting four features per kernel: adding to \textit{ppv} the \textit{mpv} (mean of positive values), the \textit{mipv} (mean of indices of positive values) and the \textit{lspv} (longest stretch of positive values). Additionally, those pooling operators are extracted for one alternative time series representation, the first order difference. This improvement comes at the cost of training time, which is approximately 2x slower, generating by default 5x more features than MiniROCKET.

\noindent \textbf{InceptionTime, a Deep Learning approach} 
The InceptionTime algorithm is one of the current most accurate DL architecture for TSC (\cite{ismail2020inceptiontime}). InceptionTime is an ensemble of five Inception networks which are using Inception-V4 modules (\cite{szegedy2017inception}) which themselves combine classic Inception blocks as introduced in \cite{szegedy2015going} and Residual Networks (ResNet) (\cite{he2016deep}).
%which was the previous top DL performer for TSC.
The InceptionTime network architecture consists in two residuals blocks, each of which containing three Inception modules,
%(see Figure \ref{fig:inception}),
it is then followed by a global averaging pooling layer and a fully-connected layer with the softmax activation function. For further information about DL for TSC, we refer to the extensive analysis of \cite{ismail2019deep}, which review the main DL algorithms for TSC.

% \begin{figure}[!htb]
%     \centering
%     \includegraphics[width=\linewidth]{figures/inception_time.png}
%     \caption{InceptionTime network architecture (\cite{ismail2020inceptiontime})}
%     \label{fig:inception}
% \end{figure}

\noindent \textbf{Hybrid/Ensemble methods} Ensemble approaches aim at combining the performance of several classifiers to built its own prediction. Those types of algorithms are currently holding the state-of-the-art in term of accuracy on the usual UEA \& UCR benchmark.\\
The Time Series Combination of Heterogeneous and Integrated Embedding Forest (TS-CHIEF) (\cite{shifaz2020ts}), the second most accurate algorithm, builds random forest of decision trees whose splitting functions are time series specific and based on similarity measures, bag-of-words representations, and interval-based transformations.\\
The Hierarchical Vote Collective of Transformation-Based Ensembles HIVE-COTE (HC~\cite{lines2018time}, HC2~\cite{middlehurst2021hive}) is the current best performer, training independently and combining several classifiers, i.e., Shapelet Transform~\cite{AB17}, Temporal Dictionary Ensemble aka TDE~\cite{MLCB20}, an ensemble of ROCKET estimators called Arsenal %(in order to compute probabilities) 
and the interval based Diverse Representation Canonical Interval Forest aka DrCIF~\cite{MLB20}.\\
Hybrid ensemble classifiers are the most accurate TSC algorithms to date. However, the top performance comes at the price of high memory and CPU resources. As a result, they are generally at least an order of magnitude slower than the other contenders (see Table~\ref{tab:time}).
\subsection{Feature Extraction tools for time series} \label{lib review}
%Most of the previously mentioned methods aim at extracting features, patterns in order to reasonably classify one given time series. In this trend, 
Besides previous algorithms dedicated to TSC problems, some research works proposed to automatically extract unsupervised features from time series in order to analyze and visualize the data. Most of them though, are rarely considered it comes to classification task and are less mentioned in TSC literature. In this part, we will briefly review the currently available packages/methods which are extracting some understandable features from time series data and which could actually be useful for classification.

%\textcolor{blue}{AB - Est-ce que ces outils sont tous non-supervisés ? Si c'est le cas, le préciser ici aussi ? (c'est dit plus bas dans 'Experimental protocol / Feature engineering tools')}

%\subsection{tsfresh} 
\noindent \textbf{Tsfresh}, which stands for Time Series FeatuRe Extraction on basis of Scalable Hypothesis tests, is a library proposing to use 63 usual times series characterization in order to extract, at most, around 800 features per sample (\cite{christ2016distributed}). It actually provides 3 pre-defined features dictionaries whose length ranging from 10 (minimal set-up, very efficient extraction) to 800 features. The package also provides some relevance filter based on some statistical tests to only keep the most informative variables among the ones you have extracted. The features computation can be parallelized, theoretically making the method more scalable. 
% Feature list : https://tsfresh.readthedocs.io/en/latest/text/list_of_features.html

%\subsection{tsfel} 
\noindent \textbf{tsfel}:The Time Series Feature Extraction Library is a more recent work which, similarly extracts a bunch of features from time series. Its specification is that it allows you to separate your data into user-specified window length and provides some better tools in order to analyse the temporal complexity of the extracted features. In tsfel, one can compute features according to their domain, the available domains are: temporal, spectral and statistical one or to extract them all at once. 
% Feature list : https://tsfel.readthedocs.io/en/latest/descriptions/feature_list.html

% \begin{figure}[!htb]
%     \centering
%     \includegraphics[width=\linewidth]{figures/tsfel_pipeline.png}
%     \caption{tsfel \textcolor{orange}{***} pipeline \cite{barandas2020tsfel}}
%     \label{fig:tsfel}
% \end{figure}

%\subsection{catch22 \& hctsa} 
\noindent \textbf{catch22 \& hctsa}: The CAnonical Time series CHaracteristics (catch22) is a subset of 22 variables from the 7730 available features in the \textit{hctsa} toolbox (\cite{fulcher2017hctsa}), which is originally a matlab package, able to extract up to nearly 7700 time series features. This limited set of features has been empirically selected over a large collection of datasets in order to choose the most explanatory ones and are, as well, minimally redundant. This approach take just a fraction of the computation time needed to compute and analyse the all \textit{hctsa} set of features. Authors exhibit the fact that concerning classification, the average performance reduction is only around $7\%$ (\cite{lubba2019catch22}).

%\subsection{tsfeatures}
\noindent \textbf{tsfeatures}: Originally a R package (\cite{hyndman2019tsfeatures}), the tsfeatures library is used to extract various type of features for time series data. The package implements around 40 methods in order to extract features, ranging from entropy to hurst exponent. A python version of the package is also available, we did not perform any time performances comparison between the two versions.

%\subsection{feasts}
\noindent \textbf{feasts}: The Feature Extraction and Statistics for Time Series (feasts) (\cite{o2019package}), looks very similar to tsfeatures. This R package provides tools for the analysis and visualization of time series. There is no python implementation of the package.

%\subsection{featuretools} 
\noindent \textbf{Featuretools} is an open-source Python library, which performs auto-features engineering based on relational tables. The library is built upon a feature discovery algorithm called \textit{Deep Feature Synthesis} (\cite{kanter2015deep}), which uses some aggregation and transform functions applied over several related tables, to create features. Working with temporal data with featuretools is quite natural, it includes many date based operations as well as some of the \textit{tsfresh} functions. While the features computation can be perform on multiple threads, the discovery though, cannot ; which make the method not very scalable when generating a large number of features. Additionally, one can remark that there is no way to prevent features generation from overfitting, generate very complex features may indeed results in learning some training set specificities and thus degrade classification performances.

%\subsection{Generalised signatures} 
\noindent \textbf{Generalised signatures}: The signature transform is an infinite collection of statistics for sequential data representation and/or feature extraction derived from rough paths theory. The signature can be thought as a moment generating function, as each term in the collection has a particular (geometrical) meaning as a function of data points. Usually, one compute the $N$ truncated signature of $x = (x_1, ..., x_n)$, with $x_i \in \mathbf{R}^d$ and linearly interpolate $x$ into the path $f=(f_1, ..., f_d):[0,1] \xrightarrow{} \mathbf{R}^d$. The signature of depth $N$ of $x$ is defined as follows: 

\begin{equation}
\footnotesize
\medmath{
    Sig^N(x) = \left(  \left( \text{  } \idotsint\limits_{0 \leq t_1 \leq ... \leq t_k \leq 1} \prod_{j=1}^k \frac{df^{i_j}}{dt} (t_j) dt_1...dt_k \right)_{1 \leq i_1,...,i_k \leq d} \right)_{1 \leq k \leq N}}
\end{equation}

The depth-1 terms ($k=1$), for example, simply represent the total increments (difference between end and start point) over each dimension. When it comes to TSC, the generalised signature pipeline (\cite{morrill2020generalised}), which has primarily been designed for multivariate series ($d > 1$), computes the signature transform over some hierarchical dyadic windows, to finally concatenate the outputs into the feature vector. This approach has shown to be competitive compared to state-of-the-art classifiers.

%% file: experiments_v2.tex
\section{Experimental Validation} \label{protocol}
In order to explore the potential of feature construction for TSC problems, the experiments carried out in this paper have been designed to answer the following questions:
\begin{itemize}
\item \textit{($Q_1$)} How do the various FC tools compare with each other in terms of predictive performance and time efficiency ?
\item \textit{($Q_2$)} In which extent using the new features perform better than raw original data ?
\item \textit{($Q_3$)} Is it beneficial to combine several feature engineering tools ?
\item \textit{($Q_4$)} How does feature construction based methods compare with state of the art TSC methods ?
\end{itemize}

Aiming at the full reproducibility of the experiments, we first present the details of our experimental protocol then discuss the obtained results. A dedicated webpage with the source code and scripts to run the experiments is also available~\footnote{\url{https://github.com/aurelien-renault/Automatic-Feature-Engineering-for-TSC}}.
% In the rest of the paper, we will be interested in reviewing \sout{some of the} libraries which generate features for time series, comparing them with themselves and eventually with some of the state-of-the-art methods from TSC literature. The main goal of this part is to put a view and some perspectives on the TSC domain, looking if some of those simple libraries alongside a wisely chosen classifier can eventually reach best competitors. In other words, is it still worth the effort of developing new complex methods only to slightly improve performances which are relatively easily attainable using simpler, sometimes more interpretable methods.  The current section will give all details about the used protocol before the results in the next section.
\subsection{Experimental protocol}
%
%\subsubsection{Datasets}

\noindent \textbf{Data sets. }We have conducted our experiments on the usual time series classification benchmark. We ran our experiments on 112 data sets out of the 128 ones which are available in the UCR Time Series Classification Archive (\cite{BLVK, dau2019ucr}). Indeed, data sets with missing values or with times series of  unequal length have been excluded from the current study, as well as the \textit{Fungi} data set which only contains one instance per class in the training set.

\begin{table*}[htbp!]
    \centering
    \begin{tabular}{lcccccc}
        \textbf{Category} & \textbf{Transformers} & \textbf{max features} & \textbf{unsup} & \textbf{embed clf} & \textbf{repr} & \texttt{sktime}\\
    \hline \hline
        \multirow{6}{6em}{\textbf{Pre-defined}} & hctsa \cite{fulcher2017hctsa} & 7730 & \textcolor{green}{\cmark} & \textcolor{red}{\xmark} & \textcolor{green}{\cmark} & \textcolor{red}{\xmark}\\
        & tsfresh \cite{christ2016distributed} & 794 & \textcolor{green}{\cmark} & \textcolor{red}{\xmark} & \textcolor{green}{\cmark} & \textcolor{green}{\cmark}\\
        & tsfel \cite{barandas2020tsfel}  & 390 & \textcolor{green}{\cmark} & \textcolor{red}{\xmark} & \textcolor{green}{\cmark} & \textcolor{red}{\xmark}\\
        & tsfeatures \cite{hyndman2019tsfeatures} & 37 & \textcolor{green}{\cmark} & \textcolor{red}{\xmark} & \textcolor{green}{\cmark} & \textcolor{red}{\xmark}\\
        & feasts \cite{o2019package} & 33 & \textcolor{green}{\cmark} & \textcolor{red}{\xmark} & \textcolor{green}{\cmark} & \textcolor{red}{\xmark} \\
        & catch22 \cite{lubba2019catch22} & 22 & \textcolor{green}{\cmark} & \textcolor{red}{\xmark} & \textcolor{green}{\cmark} & \textcolor{green}{\cmark}\\
    \hline
        \multirow{2}{6em}{\textbf{Constructed}} & featuretools \cite{kanter2015deep} & inf & \textcolor{green}{\cmark} & \textcolor{red}{\xmark} & \textcolor{green}{\cmark} & \textcolor{red}{\xmark}\\
        & fears \cite{bondu2019fears} & inf & \textcolor{red}{\xmark} & \textcolor{green}{\cmark} & \textcolor{green}{\cmark} & \textcolor{red}{\xmark}\\
    \hline
        \multirow{1}{6em}{\textbf{Signature}} & gen. signature & inf & \textcolor{green}{\cmark} & \textcolor{red}{\xmark} & \textcolor{red}{\xmark} & \textcolor{green}{\cmark}\\
    \hline
        \multirow{2}{6em}{\textbf{Convolution}} & shapelet \cite{ye2009time} & $n\_shapelets$ & \textcolor{red}{\xmark} & \textcolor{red}{\xmark} & \textcolor{red}{\xmark} & \textcolor{green}{\cmark}\\
        & rocket \cite{dempster2020rocket} & inf & \textcolor{green}{\cmark} & \textcolor{red}{\xmark} & \textcolor{green}{\cmark} & \textcolor{green}{\cmark}\\
    \hline
        \multirow{2}{6em}{\textbf{Symbolic (SFA)}} & BOSS \& cie (\cite{schafer2015boss}, \cite{large2019time}, \cite{middlehurst2019scalable}) & $c^l$ & \textcolor{red}{\xmark} & \textcolor{red}{\xmark} & \textcolor{red}{\xmark} & \textcolor{green}{\cmark}\\
        & WEASEL \cite{schafer2017fast} & $c^{2l}$ & \textcolor{red}{\xmark} & \textcolor{red}{\xmark} & \textcolor{red}{\xmark} & \textcolor{green}{\cmark}\\
    \hdashline
        \textbf{(SAX)} & BOP \cite{lin2003symbolic} & $c^l$ & \textcolor{green}{\cmark} & \textcolor{red}{\xmark} & \textcolor{red}{\xmark} & \textcolor{green}{\cmark}\\ 
    \hline
        \multirow{2}{6em}{\textbf{Intervals}} & TSF \cite{deng2013time} & $3r\sqrt{m}$ & \textcolor{red}{\xmark} & \textcolor{green}{\cmark} & \textcolor{red}{\xmark} & \textcolor{green}{\cmark}\\
        & CIF \cite{middlehurst2020canonical} & $25r\sqrt{m}$ & \textcolor{red}{\xmark} & \textcolor{green}{\cmark} & \textcolor{red}{\xmark} & \textcolor{green}{\cmark}\\
    \hline
        \multirow{2}{6em}{\textbf{Deep Learning}} & AE/VAE \cite{kingma2013auto} & \multirow{2}{6em}{$latent\_dim$} & \textcolor{green}{\cmark} & \textcolor{red}{\xmark} & \textcolor{red}{\xmark} & \textcolor{red}{\xmark}\\
        & TS2Vec \cite{yue2022ts2vec} & & \textcolor{green}{\cmark} & \textcolor{red}{\xmark} & \textcolor{red}{\xmark} & \textcolor{red}{\xmark}\\
    \end{tabular}
    \caption{ Comparison table of several time series transformation methods. (max features: maximum number of generated features, unsup: unsupervised features extraction, embedded clf: use a specific classifier, i.e. can not be used with any off-the-shelf classifier, repr: use of different TS representations, - sktime: present in sktime framework (0.13.0)}
    \label{tab:comp}
\end{table*}

%\subsubsection{Tested libraries}
\noindent \textbf{Feature engineering tools. }
As one of the main goal is to compare several feature extraction methods, it is more suitable to separate the feature generation step from the classification one. Thus, we decided to include in this study only the unsupervised methods, which can be easily used with any classifier (see Table \ref{tab:comp}). We removed deep learning approaches from our study, since generative models like AutoEncoders require very high CPU resources -- which leads to hardly fair comparison -- and, as highlighted in~\cite{ismail2019deep}, their predictive performance on TSC problems is mitigated. In addition, based on the TSF, CIF algorithms, as in \cite{middlehurst2022freshprince}, we have included two methods for extracting some set of descriptive statistics on random sampled intervals i.e. subseries. Nevertheless, since we want to be able to use any off-the-shelf classifier, we are extracting the whole set of features on all the sampled intervals rather than learning one Decision tree on some features subsets as is TSF, CIF.
%
%Moreover, no Deep Learning approaches have been studied, as highlighted in \cite{ismail2019deep}, generative models like Auto-Encoder are usually tacitly considered to be less accurate than direct discriminate ones, models which are not really in our feature generation / representation learning scope. Deep Learning techniques cannot run on CPU which, as well, make the comparison hardly fair.

Finally, we tested a large panel of libraries which are the following: 

\begin{itemize}
\item ROCKET and its variants (\cite{dempster2020rocket}, \cite{dempster2021minirocket}, \cite{tan2021multirocket}),
\item Intervals Transform (which simply apply a summary transformation over multiple random intervals)
\item Intervals Catch22 (\textit{catch22} features from random intervals)
\item TsFreh (\cite{christ2016distributed}),
\item Tsfel (\cite{barandas2020tsfel}),
\item Catch22 (\cite{lubba2019catch22}),
\item TsFeatures (\cite{hyndman2019tsfeatures}),
\item FeatureTools (\cite{kanter2015deep}),
\item Signature (\cite{morrill2020generalised})
\item Feasts (\cite{o2019package}),
\item HCTSA (\cite{fulcher2017hctsa}).
\end{itemize}

% TSFRESH~\cite{christ2016distributed,CBN+18}, TSFEL~\cite{barandas2020tsfel}; and R-based  TSFEATURES~\cite{hyndman2019tsfeatures}, FEASTS~\cite{o2019package}. Originally designed for descriptive analysis and visualization of time series, their use for TSC problems has been less studied: e.g., B.~Fulcher et al.~\cite{fulcher2017hctsa} suggest a linear classifier based on the feature set generated by HCTSA and on a much smaller set, CATCH22~\cite{lubba2019catch22}. 

%\subsubsection{Experimental parameters}
\noindent \textbf{Experimental parameters and classifiers. }
In order to keep the comparison as fair as possible, we evaluated the different methods limiting the number of features to 1000.  Of course, not all libraries generate this number of extracted features (see Table \ref{tab:comp}), \textit{tsfresh} for example, cannot generates more than 794 features without explicitly telling it what to compute. When available, the transformations algorithms have been run using the \texttt{sktime} package (\cite{loning2019sktime}). Hyperparameters values are reported in Table \ref{tab:hyp}. All experiments have been computed using a single thread of a 4-core 2GHz Intel Core i5-1038NG7 CPU, except for \textit{hctsa} which used 4 workers, with 16Go RAM. Once all the features matrices have been computed, we applied 9 different classifiers, Random Forests (100, 500 trees), Logistic Regression ($\ell_{2}$, $elasticnet$ penalties), XGBoost, SVM (linear, $rbf$ kernel), 1 Nearest Neighbors and Rotation Forest with every other parameters being the default ones, from the \texttt{sklearn} library (\cite{pedregosa2011scikit}), except for the Rotation Forest classifier, which is in the \texttt{sktime} package~\cite{bagnall2018rotation}.

% \textcolor{magenta}{parties ci-dessous reprises des annexes du stage}

% Table \ref{tab:hyp} displays hyper-parameters for the tested features extraction libraries. For the comparison to be fair, each considered method has been to the number of generated features the closer to 1000. This number could be obtained through several parameters combinations, though, we do not seek to optimize these and select the most straightforward way to reach/limit this arbitrarily selected number of features.  \textcolor{magenta}{The classifiers' paramaters are mainly those by default and could be found in  the supplementary material as well as additional details}.
%MAIS y-a-t-il cross val sur les params des classifieurs, l'annexe n'est pas si claire ? \textcolor{violet}{non pas de cross-val sur aucun params : ni lib, ni classifiers. Parfois différentes valeurs de params pour un même clf, ex : ajout ou non l1 penalty ds une LogisticRegression}
 
\begin{table}[!htb]
    \centering
    \renewcommand{\arraystretch}{1.2}
    \setlength{\tabcolsep}{1pt}
    \begin{tabular}{|c|c|c|}
    \hline
          \textbf{Methods} & \textbf{Parameters} & \textbf{Nb features} \\
    \hline
          rocket & \footnotesize $n\_kernels: 500$ & 1000 \\
    \hline
          minirocket &  \footnotesize $n\_kernels: 1000$ & 924 \\
    \hline
          \multirow{2}{5em}{multirocket} &  \footnotesize $n\_kernels: 125$ & \multirow{2}{2em}{1008} \\
          &  \footnotesize $n\_features\_per\_kernel: 6$ & \\
    \hline
        \multirow{3}{5em}{intervals} &  \footnotesize $n\_intervals: 100$ & \multirow{3}{2em}{1000} \\
        & \footnotesize $agg: [mean, min, max, sum, med, std$ & \\
        & \footnotesize $count, skew, quant(0.25), quant(0.75)]$ & \\
    \hline
        intervals c22 & \footnotesize $n\_intervals = 45$ & 1000 \\
    \hline
        \multirow{4}{5em}{featuretools} & \footnotesize $agg: [mean, min, max, sum, $& \multirow{4}{1em}{50} \\
        & \footnotesize \footnotesize $ median, std, count, skew]$ & \\
        & \footnotesize $transf: [D, DD, CUMSUM,$ & \\
        & \footnotesize $DCUMSUM, ACF, PS]$ & \\
    \hdashline
      featuretools\_1k & \footnotesize $col\_combinations : [add, substr, mult]$ & 1000 \\
    \hline
        \multirow{2}{4em}{signature} & \footnotesize $window\_name: dyadic$ & \multirow{2}{1em}{930} \\
        & \footnotesize $window\_depth: 4$ & \\
    \hline
        hctsa & \footnotesize $n\_features : 1000$ & 1000 \\
    \hline
        catch22 & \multirow{5}{10em}{\footnotesize $default \;\; parameters$} & 22 \\
        feasts & & 33 \\
        tsfeatures & & 37 \\
        tsfel & & 142-390 \\
        tsfresh & & 789 \\
    \hline
    \end{tabular}
    \caption{Libraries' hyperparameters values}
    \label{tab:hyp}
\end{table}

%\subsubsection{Evaluation methodology}
\noindent \textbf{Evaluation methodology. } We use pre-defined train/test sets available from the UCR archive and the accuracy as the predictive performance evaluation measure. This way, the comparison with state of the art methods like HC1, HC2, TSChief and InceptionTime -- which are computationally expensive -- is feasible as we report the results from the original papers.
To compare different approaches over several data sets, critical difference diagrams have been drawn using the post-hoc Nemenyi rank based test. Although \cite{benavoli2016should} suggests to use the Wilcoxon signed-rank test with the Holm correction in place of the Nemenyi one, the diagrams drawn this way have been found to be unreadable quite often as it contained some overlapping cliques, as the Wilcoxon signed-rank test is not based on mean ranks, the analysis could have been confusing. Though, we display, when needed, alongside critical Nemenyi diagrams, the binary matrix showing explicitly the Holm adjusted pairwise Wilcoxon comparison for each considered methods (see subpart (b) of figures). In these charts, the small black squares identify pairs of approaches that do not differ significantly in performance. %The performance metric used further is the accuracy score. Further results, including many others metrics, are available in supplementary material.
 
%%%%%%%%%%%%%%%%%%%%%%%%%%%
\subsection{Experimental results and Discussion}
\label{results}

%\textcolor{magenta}{petit texte d'intro à ajouter,  anoncer axes d'analyse (perf, time, interprtability)...}

\begin{figure}[!th]
%\fbox{
%\parbox[c]{0.95\linewidth}{
\centering
\subfloat[Critical diagram labeled with mean accuracy]{\includegraphics[width=0.8\linewidth]{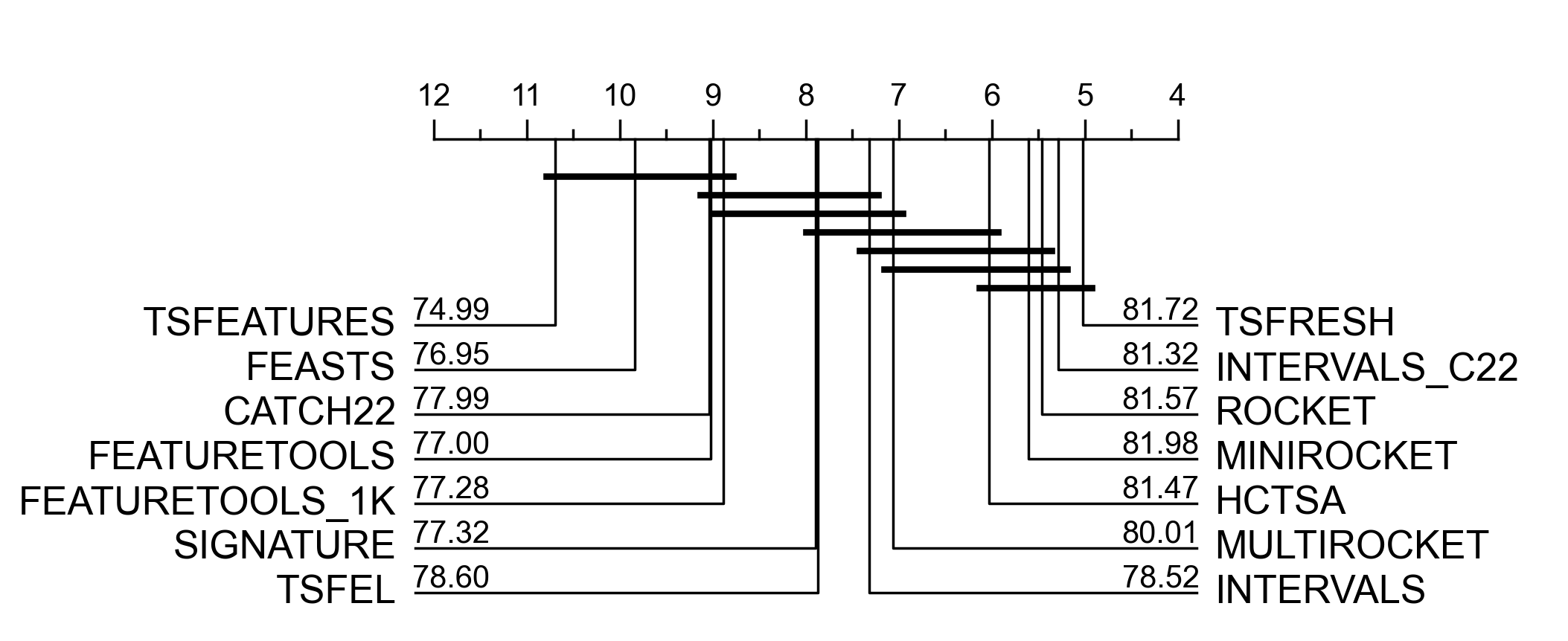}}\\
\subfloat[Corrected pairwise comparison]{\includegraphics[width=0.55\linewidth]{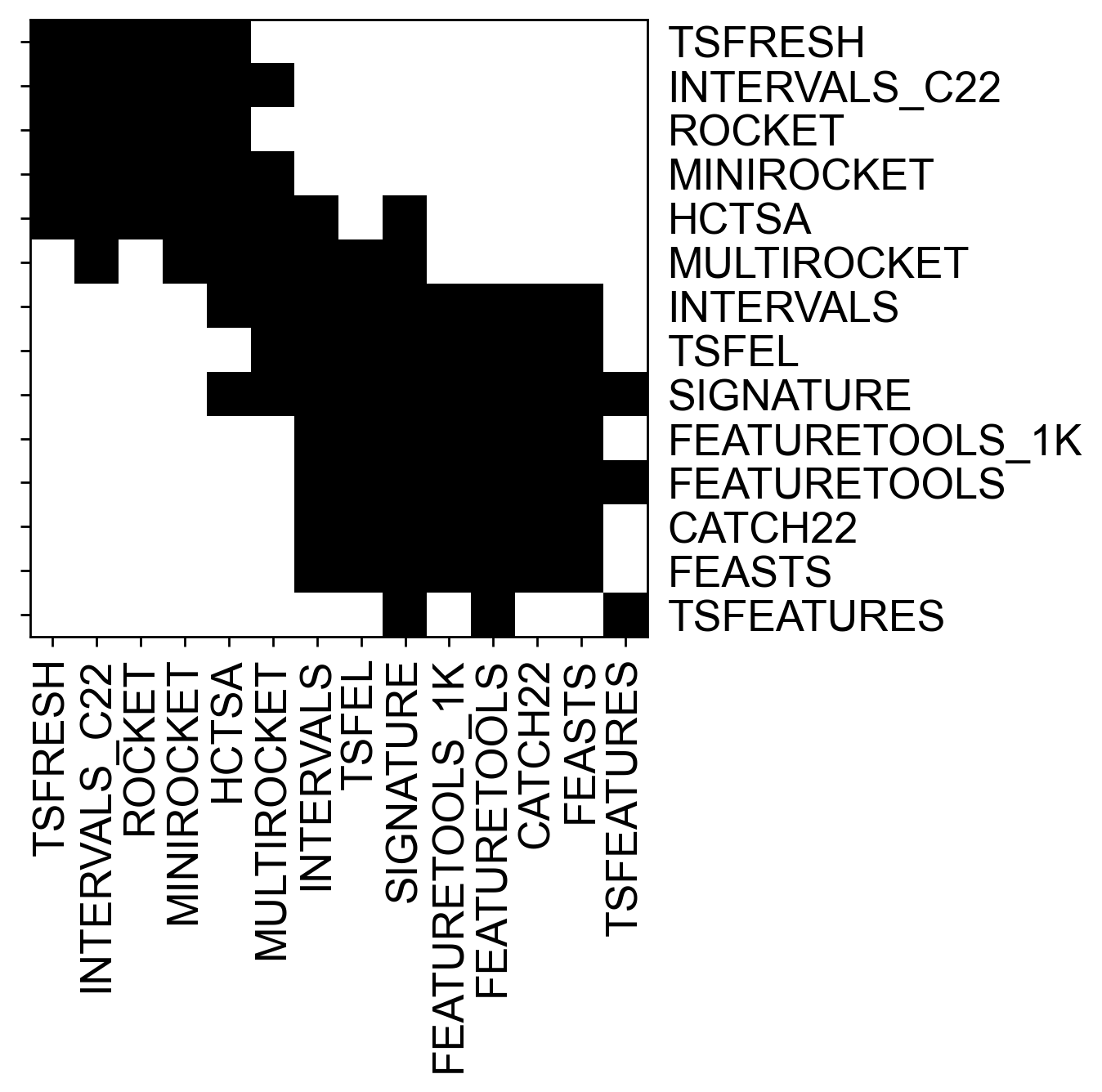}}
%}}
\caption{Random Forest}
\label{fig:acc_rf}
\end{figure}

%\subsection{Performances} 
\noindent \textbf{Performance comparison of feature engineering tools. }
Due to space limitation, we only report charts for three classifiers: Random Forests, Rotation Forests and Logisitc Regression.\\
For a default Random Forest classifier, the best performing methods are the ROCKET-like methods, \textit{tsfresh}, \textit{hctsa} and the random intervals extracting the \textit{catch22} features.(see figure \ref{fig:acc_rf}). 

\begin{figure}[!th]
\centering
\subfloat[Critical diagram labeled with mean accuracy]{\includegraphics[width=0.8\linewidth]{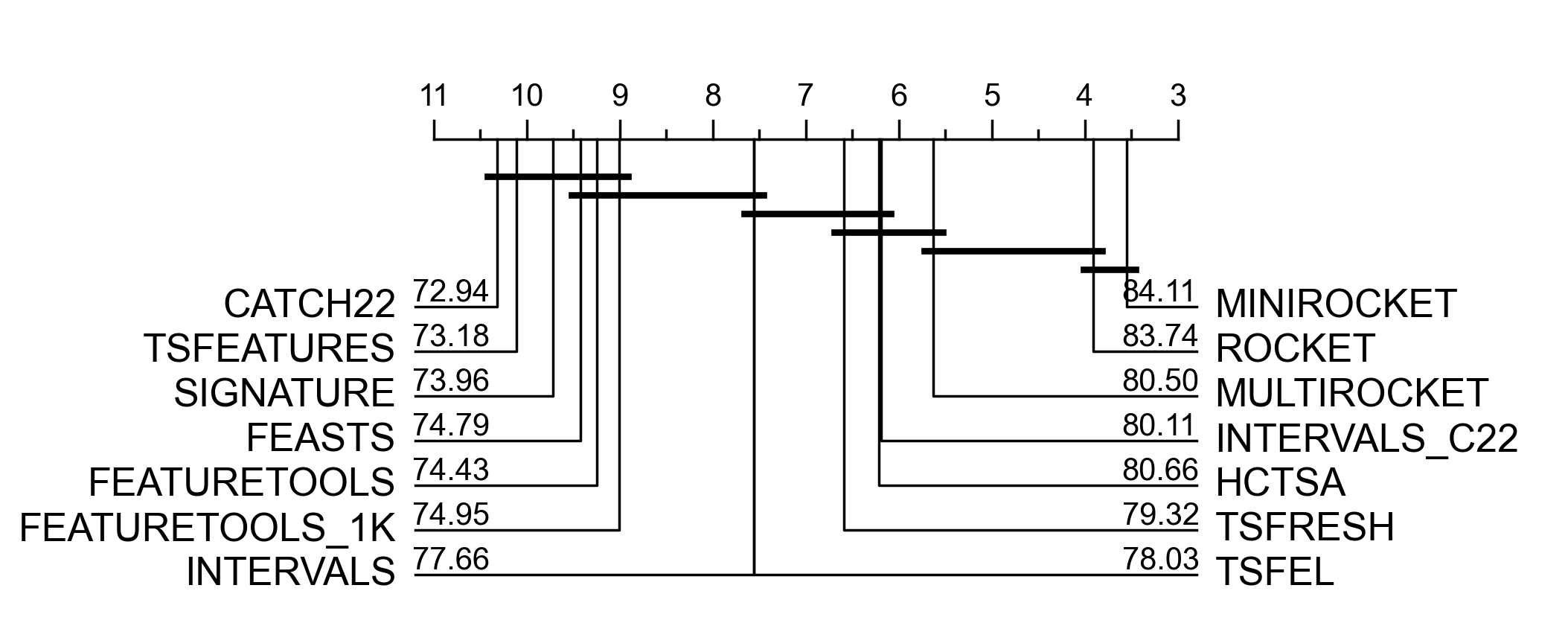}}\\
\subfloat[Corrected pairwise comparison]{\includegraphics[width=0.5\linewidth]{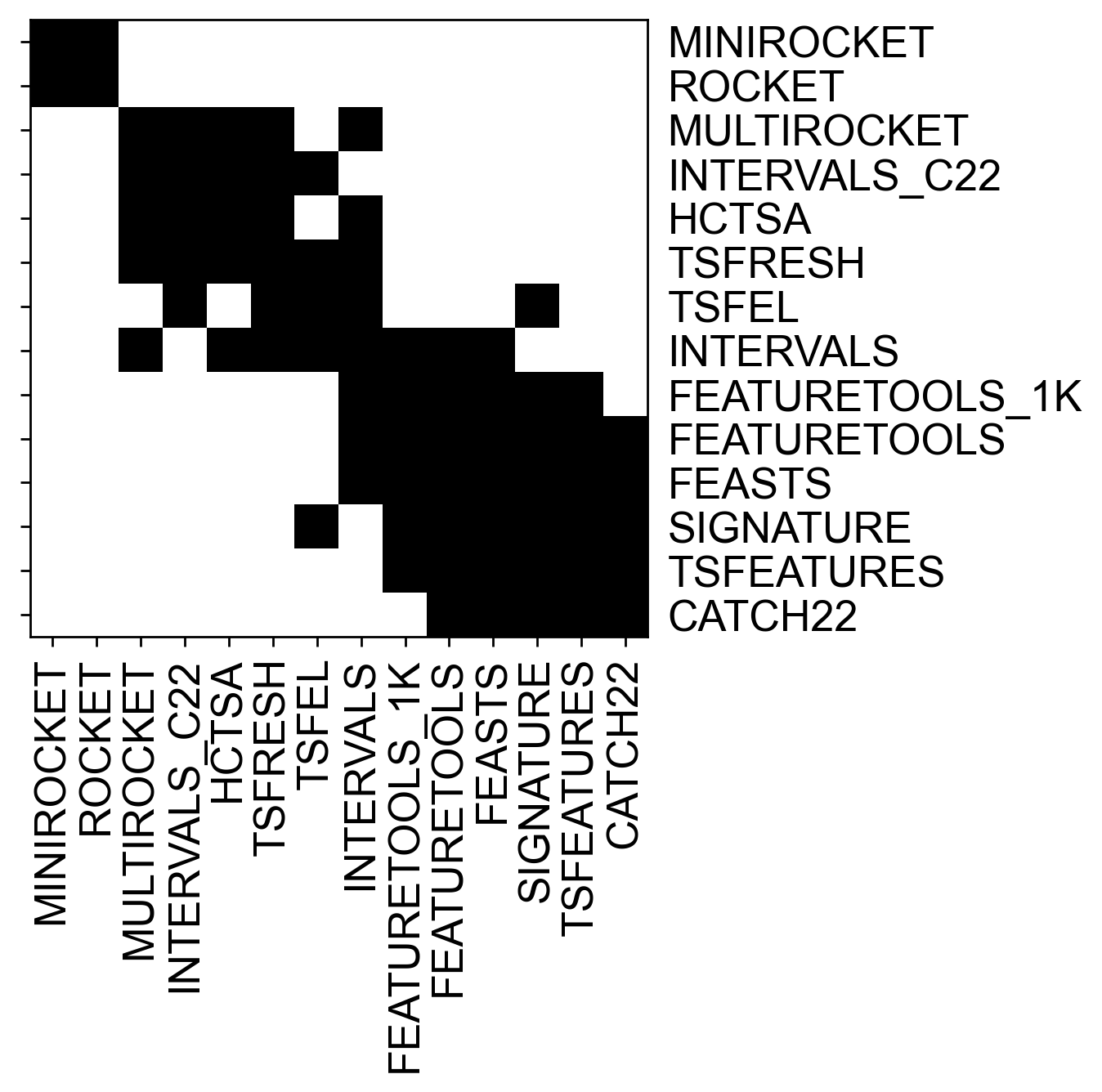}}
\caption{Logistic Regression }
\label{fig:acc_log}
\end{figure}

When dealing with linear classifiers though, the ROCKET-like approaches become significantly better than the other tested methods as they are likely to generate independent kernels (see Figure \ref{fig:acc_log}). 
On the other side, the \textit{tsfresh}/\textit{hctsa} features are more likely to be correlated, which can partially explain that best classifiers are non linear ones such as Random Forest or XGBoost. It's also worth noting that data is standardized before being processed by linear classifiers, as they usually embed penalization, which make some of the extracted features useless.

\begin{figure}[!th]
\centering
\subfloat[Critical diagram labeled with mean accuracy]{\includegraphics[width=0.8\linewidth]{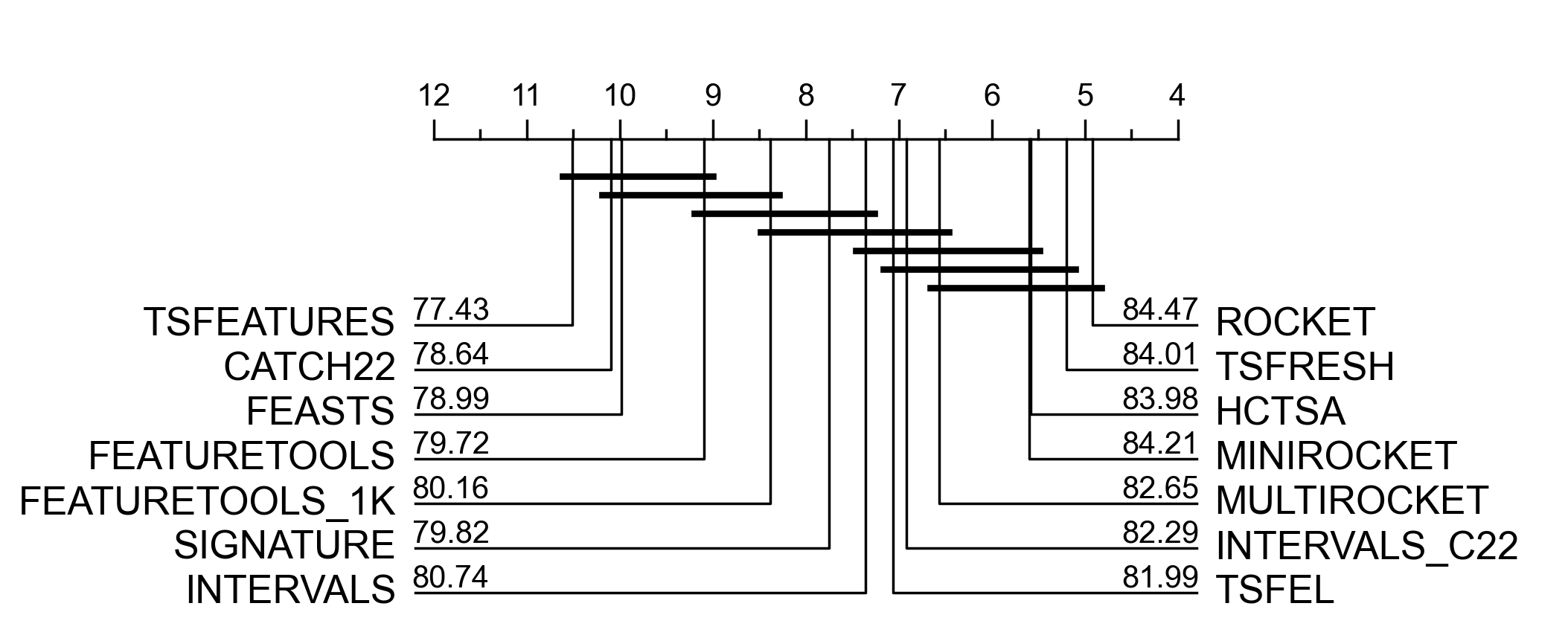}}\\
\subfloat[Corrected pairwise comparison]{\includegraphics[width=0.5\linewidth]{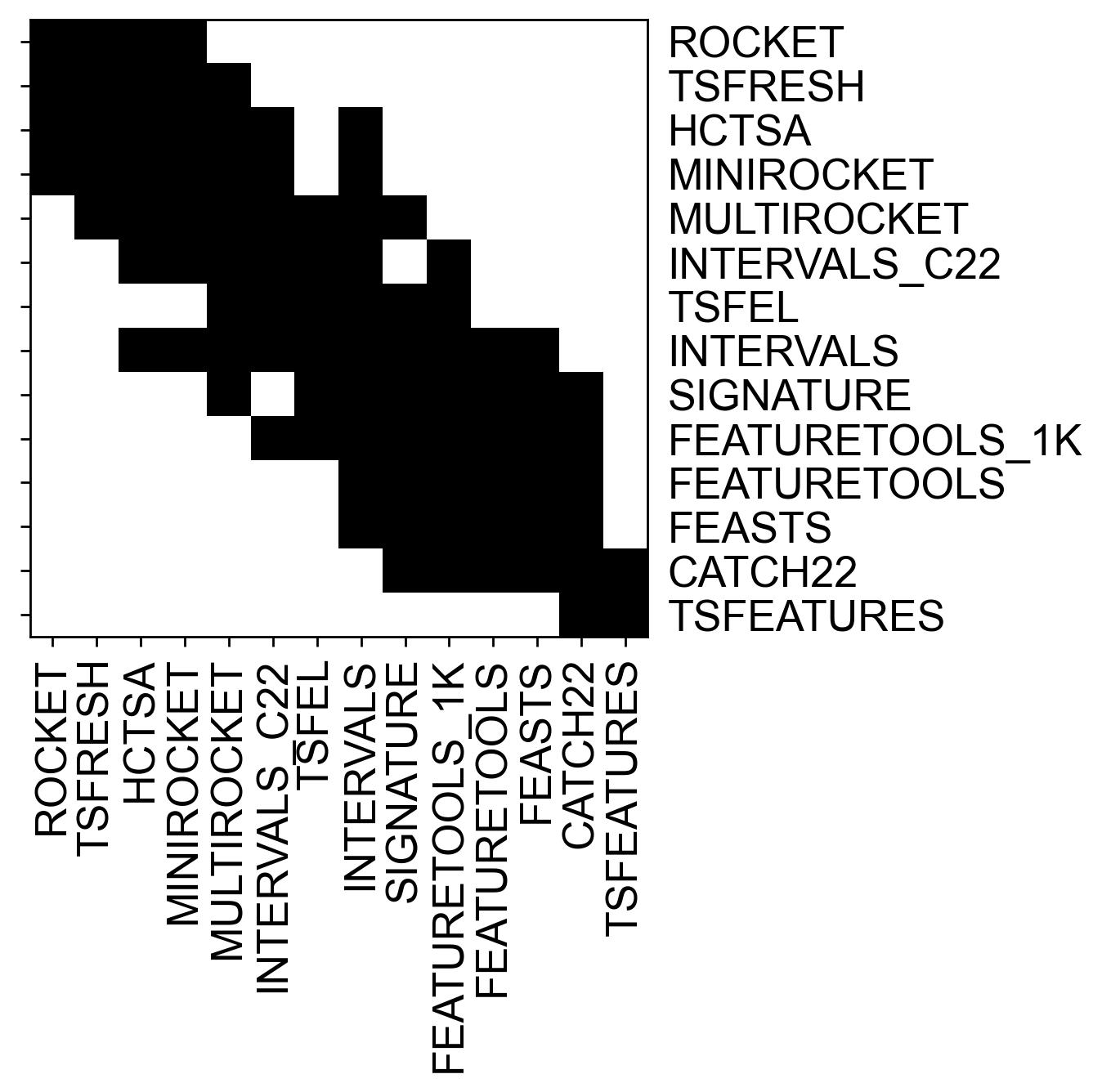}}
\caption{Rotation Forest }
\label{fig:acc_rotf}
\end{figure}

We are also displaying the results obtained for the Rotation Forest classifier \cite{lin2012rotation} (see Figure \ref{fig:acc_rotf}), which is actually the best performer for the great majority of the tested libraries, even if generally less scalable. %Those observations are strengthened when running our experiments with a all bunch of different classifiers, the different tested classifiers alongside their parameters can be found in the supplementary material \textcolor{magenta}{(see XXXX).}

\begin{figure}[!th]
\centering
\subfloat[Critical diagram labeled with mean run time in seconds]{\includegraphics[width=1.0\linewidth]{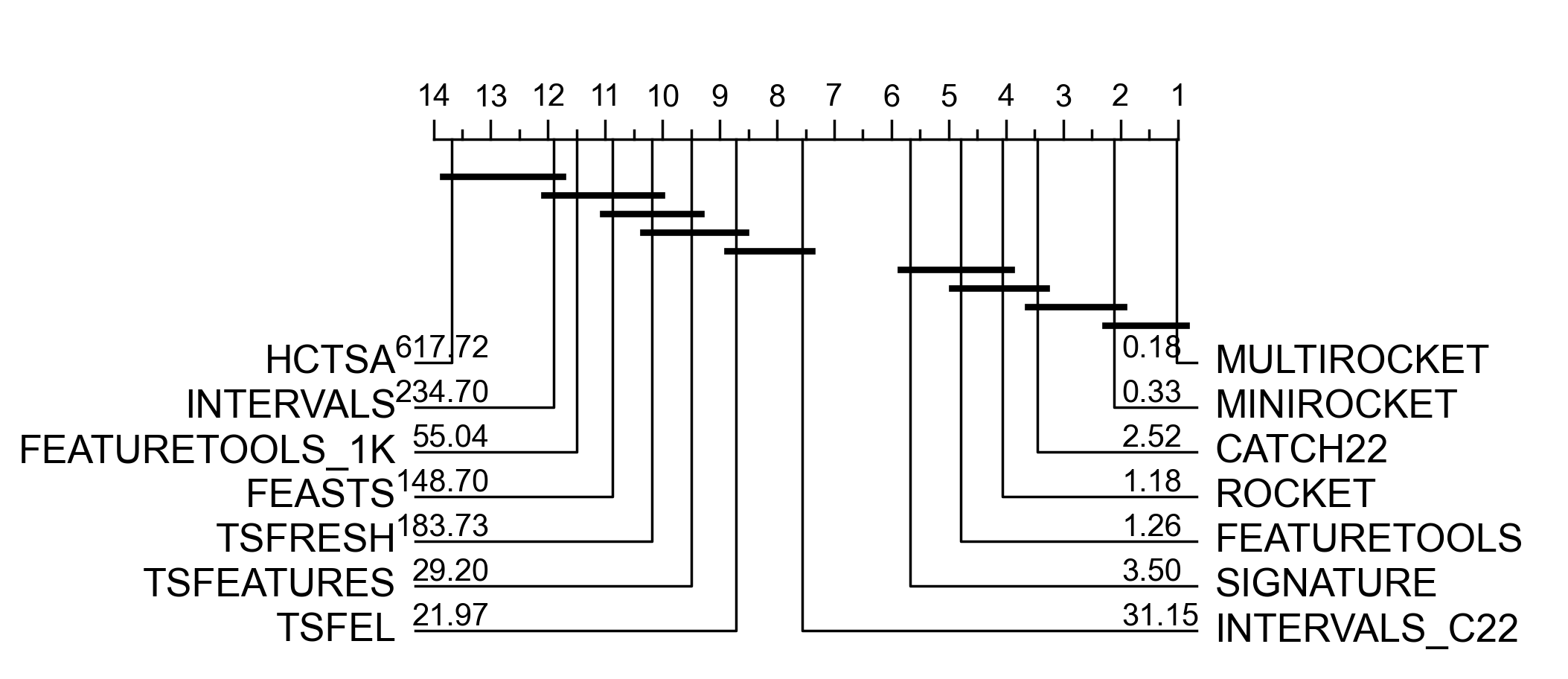}\label{fig:rt}}\\
\subfloat[Critical diagram labeled with mean time/feature in seconds per feature]{\includegraphics[width=1.0\linewidth]{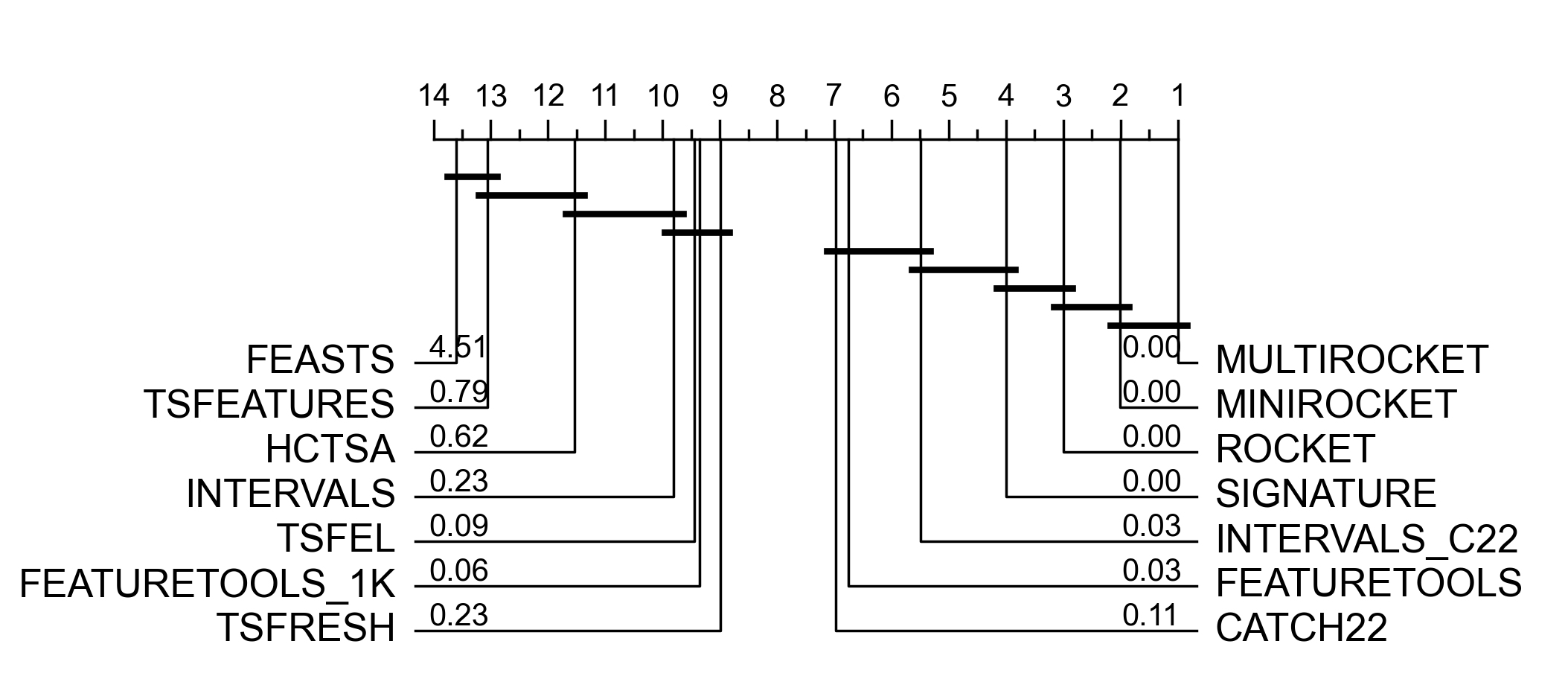}\label{fig:tf}}
\caption{Computation time performances}
%\label{fig:tf}
\end{figure}

%\subsection{Computation time and complexity}
\noindent \textbf{Running times comparison. }
When considering computation time, we observe a clear trend: as expected, all the ROCKET-features based methods  outperform the others in term of both time per feature and overall run time. %, as those are very optimised to compile fast as mentioned in \ref{literature review}. 
In average the ROCKET-like methods take less than half a second for one given data set. Among the less scalable methods, we find libraries as \textit{hctsa}, \textit{feasts} or \textit{tsfeatures}, as those are designed for analysis and visualization rather than large scale extraction to be include in a automatic classification pipeline. \textit{Hctsa} average run time, for example, is around 10 minutes for one data set (see Figure \ref{fig:rt}), \textit{feasts} take more than 4 seconds to extract one feature, which make them the worst scalable libraries among the tested methods. Furthermore, one can notice that bringing \textit{featuretools} from 50 to 1000 generated features slow down the algorithm, doubling the time/feature metric (see Figure \ref{fig:tf}), which validate the previously mentioned hypothesis made about the library scalability.

\subsection{Raw data, feature construction or both} 

\begin{figure}[!th]
\centering
\subfloat[Strategies comparison: raw data (RAW), with Feature (FTS) and RAW+FTS, for all classifiers]{\includegraphics[width=1.0\linewidth]{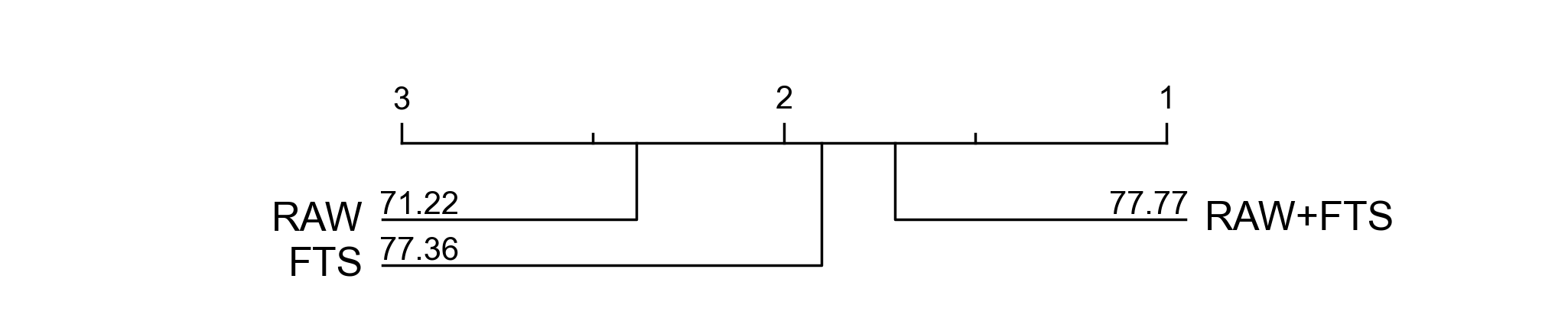}\label{fig::raw:a}}\\
\subfloat[Strategies comparison for minirocket feature, for all classifiers]{\includegraphics[width=1.0\linewidth]{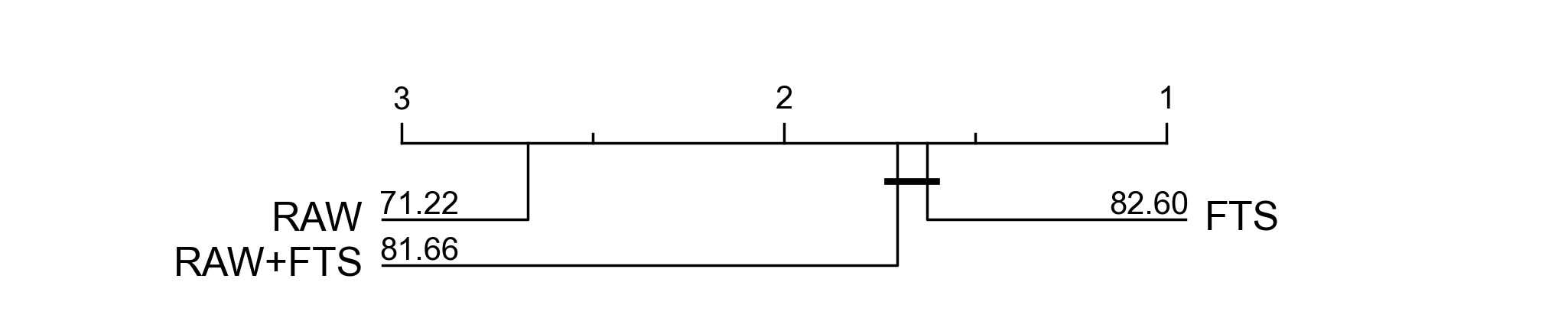}\label{fig::raw:b}}
\caption{Performance comparison: adding the raw time series or not labeled with mean accuracy}
\label{fig::raw}
\end{figure}

In order to put some more perspective on our results we analyze the performance when using only the raw time series, i.e. each time point is considered as one feature, as well as the concatenation of the raw data with the extracted features and compare those with the previously obtained results, only using the computed features. Considering these three strategies, the first conclusion is that, extracting features, no matter which method is used, outperforms practically every time a classification only based on the raw time series. As well, Figure \ref{fig::raw} demonstrates that, on average, across all strategies for all classifiers, adding the raw data to the feature matrix is performing significantly better. 

\begin{figure}[!th]
\centering
\subfloat[$ts\_length < 315$]{\includegraphics[width=1.0\linewidth]{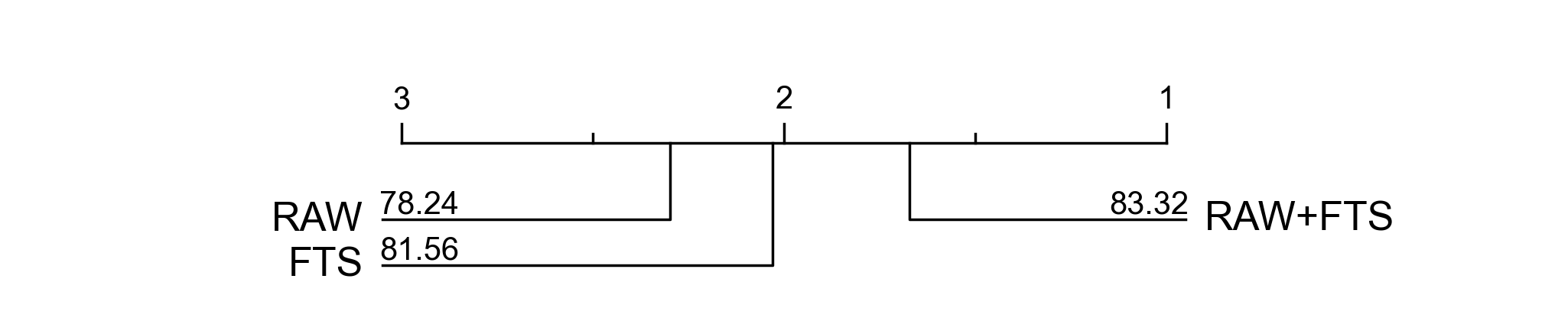}\label{fig::ts_len:a}}\\
\subfloat[$ts\_length > 315$]{\includegraphics[width=1.0\linewidth]{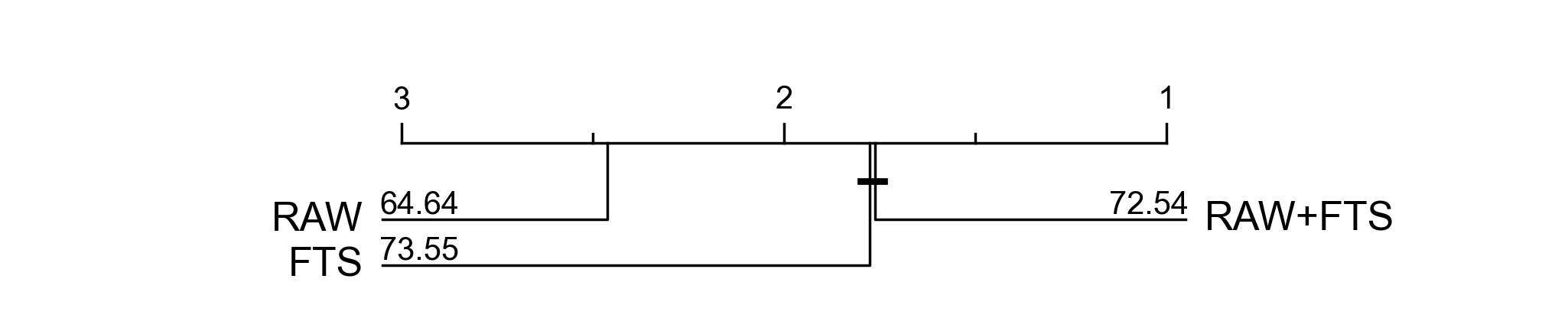}\label{fig::ts_len:b}}\\
\subfloat[$ts\_length > 720$]{\includegraphics[width=1.0\linewidth]{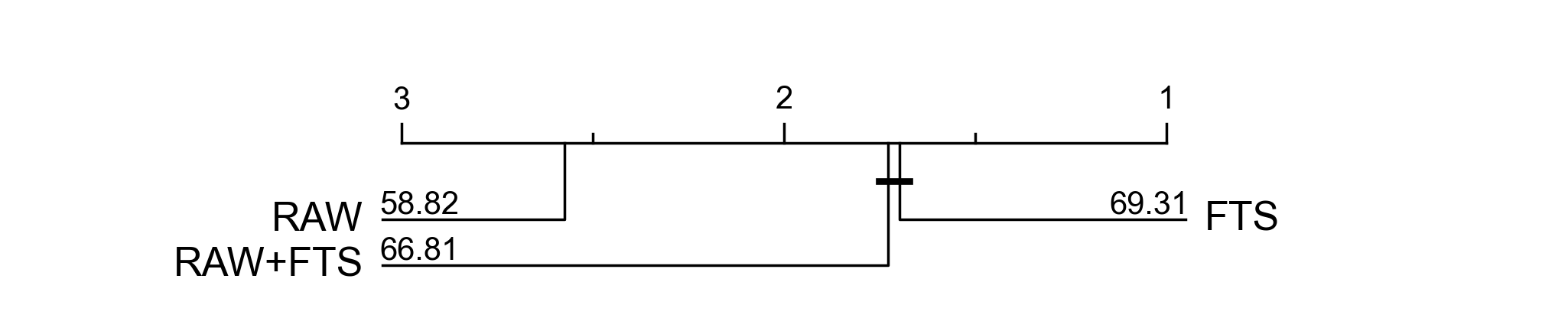}\label{fig::ts_len:c}}
\caption{Accuracy results for all strategies on all classifiers considering time series length}
\label{fig::ts_len}
\end{figure}

A more in-depth analysis actually shows that this improvement can be explained by the libraries which tend to generate correlated features. Indeed, in those cases, the best classifiers are the ones that are less sensitive to feature correlation and redundancy, i.e. Random Forest or XGBoost; adding the values of the time series (which could be correlated), when the length of the considered time series remains under a certain threshold, is improving performance in a significant way. However, as the time series length grows, this adding tends to add more noise and, at some point, harm performances, in Figure \ref{fig::ts_len}, we are displaying the performance only considering data sets for which the length of the time series are below/above the median length of the UCR repository.

Additionally, one can notice that, only working with data sets whose time series length are beyond the third quartile (around 28 datasets), the overall performance is much degraded and adding the raw data is then ranked behind the FTS strategy. The methods for which linear classifiers were the best never benefit from the raw data: even if some classifiers handle this incorporation better (adding a $\ell_1$ penalty for example), they are able, at best, not to degrade the performance too much.

%\subsubsection{Libraries complementarity}

\subsection{Combining libraries} 

In this part, we define one strategy in order to figure out if performance can be improved by combining several feature engineering tools. To keep things simple, the tested approach consists in gradually stacking the different methods, ranked by one given metric, here accuracy, obtained with a vanilla Random Forest classifier. We are not considering the whole set of ROCKET-like methods any longer, we decided to only retain the 1000 features created by MiniROCKET in a fisrt experiment, before completely deleting MiniROCKET features in a second one. Indeed, as ROCKET represents, a TSC approach on its own, we wanted to address whether or not our features set could still be competitive not using this great performing classifier. Additionally, to reduce computation time and redundancy, the expanded version of \textit{featuretools} (with 1000 features) is also removed, as it did not provide any more significant information than its lighter version (with 50 features). 

\begin{figure}[!th]
\centering
\subfloat[Rotation Forest, the best ranked strategy is called: \textcolor{red}{Features}]{\includegraphics[width=1.0\linewidth]{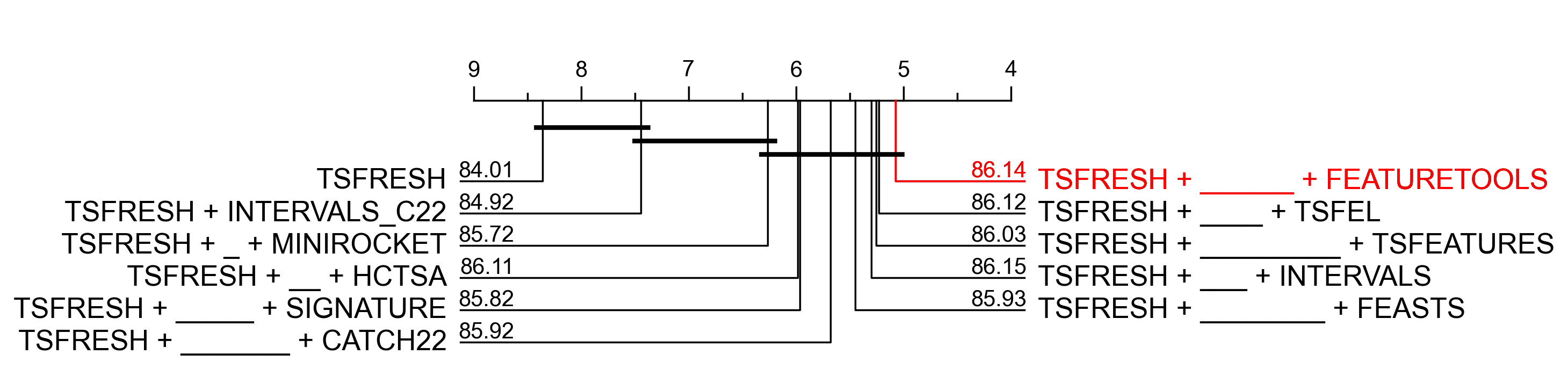}\label{fig:rfs}}\\
\subfloat[Rotation Forest, without ROCKET methods, the best ranked strategy is called: \textcolor{red}{Features\_noROCKET}]{\includegraphics[width=1.0\linewidth]{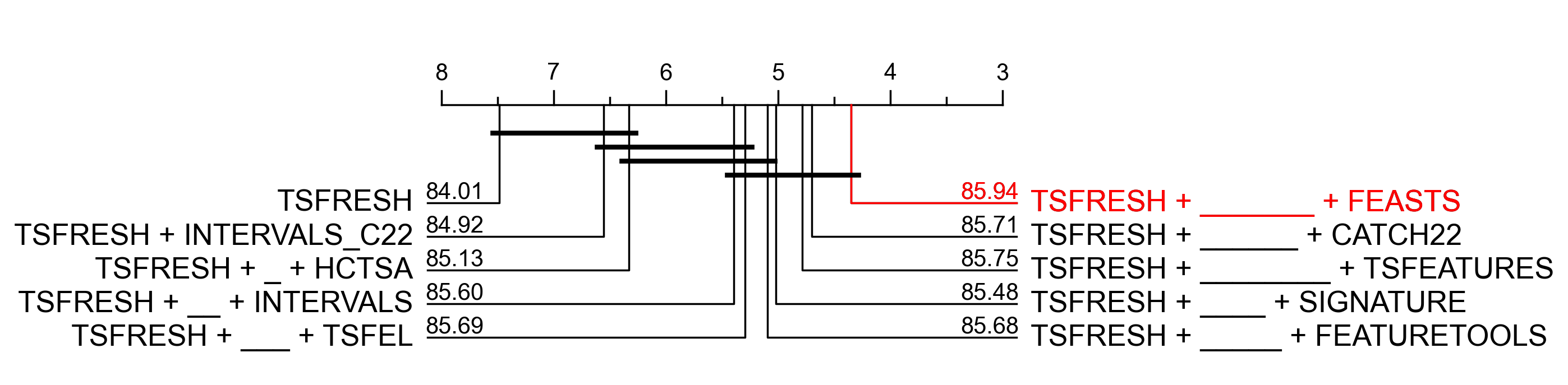}\label{fig:rotfs}}
\caption{Critical diagrams labeled with mean accuracy \\  \footnotesize Only the first and last library are displayed, one underscore stands for one in-between library like in the following : \\ TSFRESH + \_ + HCTSA = TSFRESH + INTERVALS\_C22 + HCTSA.}
\end{figure}
% expliquer stacking (underscore == another lib)

It seems intuitive that stacking the different features ones onto others will probably add redundancy; in this way classifier like random/rotation forest seem to better fit this purpose (\cite{bagnall2018rotation}), we only reports the results for the Rotation Forest in what follows as it provided best performance so far. For this classifier, the best stacking strategy both in terms of ranks and mean accuracy (see Figure \ref{fig:rfs}) is the ones using 8 tools from the all library set, excluding the ones which are not performing well on their own i.e. \textit{catch22}, \textit{feasts} and \textit{tsfeatures}, which are also the ones creating a limited number of features. When removing MiniROCKET features, some of those libraries become informative and the best strategy, without MiniROCKET, is now containing the all libraries set, only excluding \textit{tsfeatures} (see Figure \ref{fig:rotfs}). The top performers of both of these experiments are called \textit{Features} and \textit{Features\_noROCKET} respectively in what follows. Those strategies though, are not performing significantly better that stacking the two or three best libraries ones onto others.

\subsection{Comparison with state of the art TSC approaches} 

\begin{figure}[htbp!]
\centering
\subfloat[Critical diagram labeled with mean accuracy]{\includegraphics[width=1.0\linewidth]{images/cd-diagram-nemenyi-Accuracy.png}\label{fig:sota-a}}\\
\subfloat[Corrected pairwise comparison]{\includegraphics[width=0.5\linewidth]{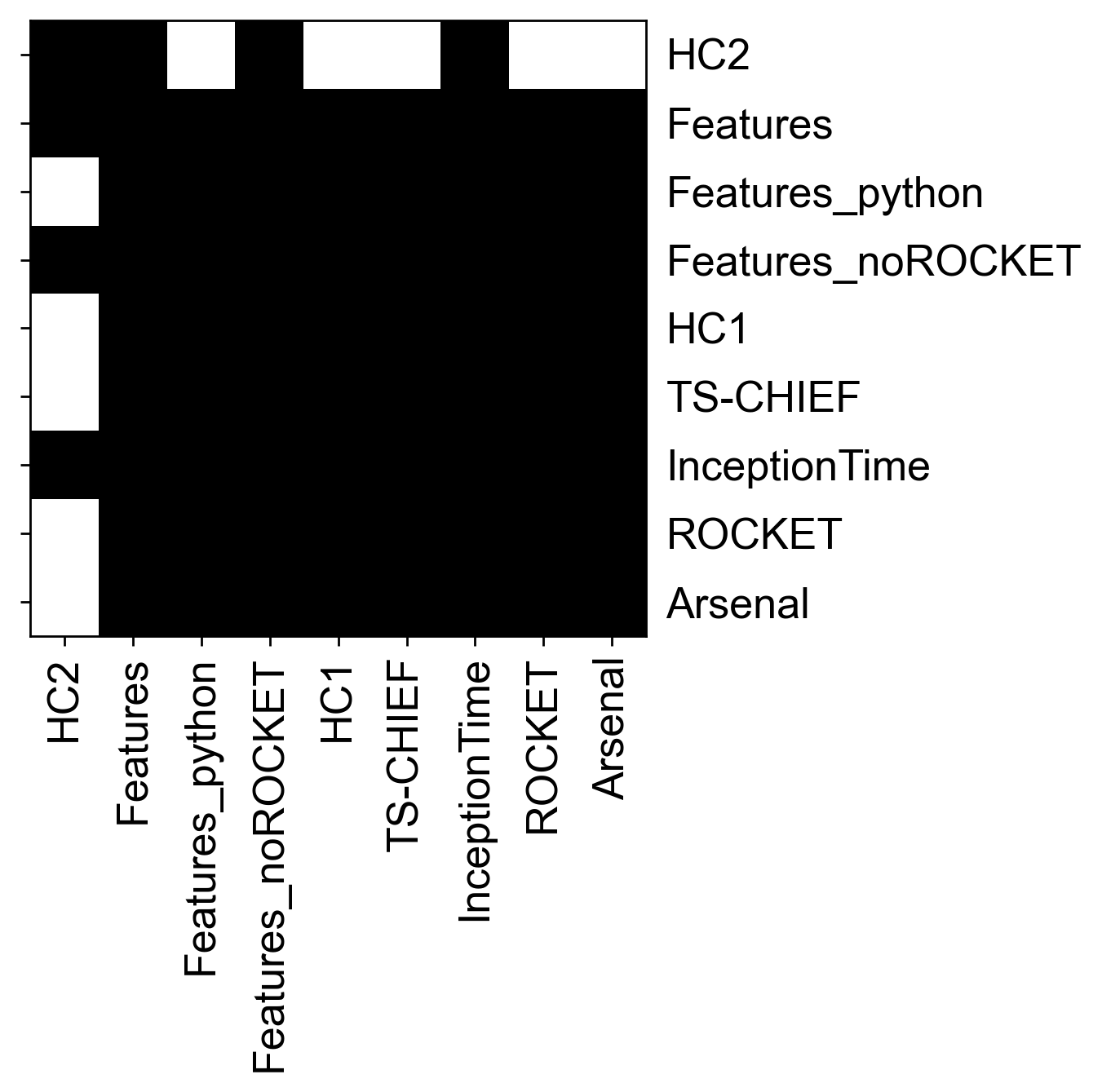}\label{fig:sota-b}}
\caption{Predictive performance comparison with state-of-the-art TSC algorithms}
\label{fig:sota}
\end{figure}

\begin{figure*}[htbp!]
    \centering
    \subfloat{\includegraphics[width=7cm]{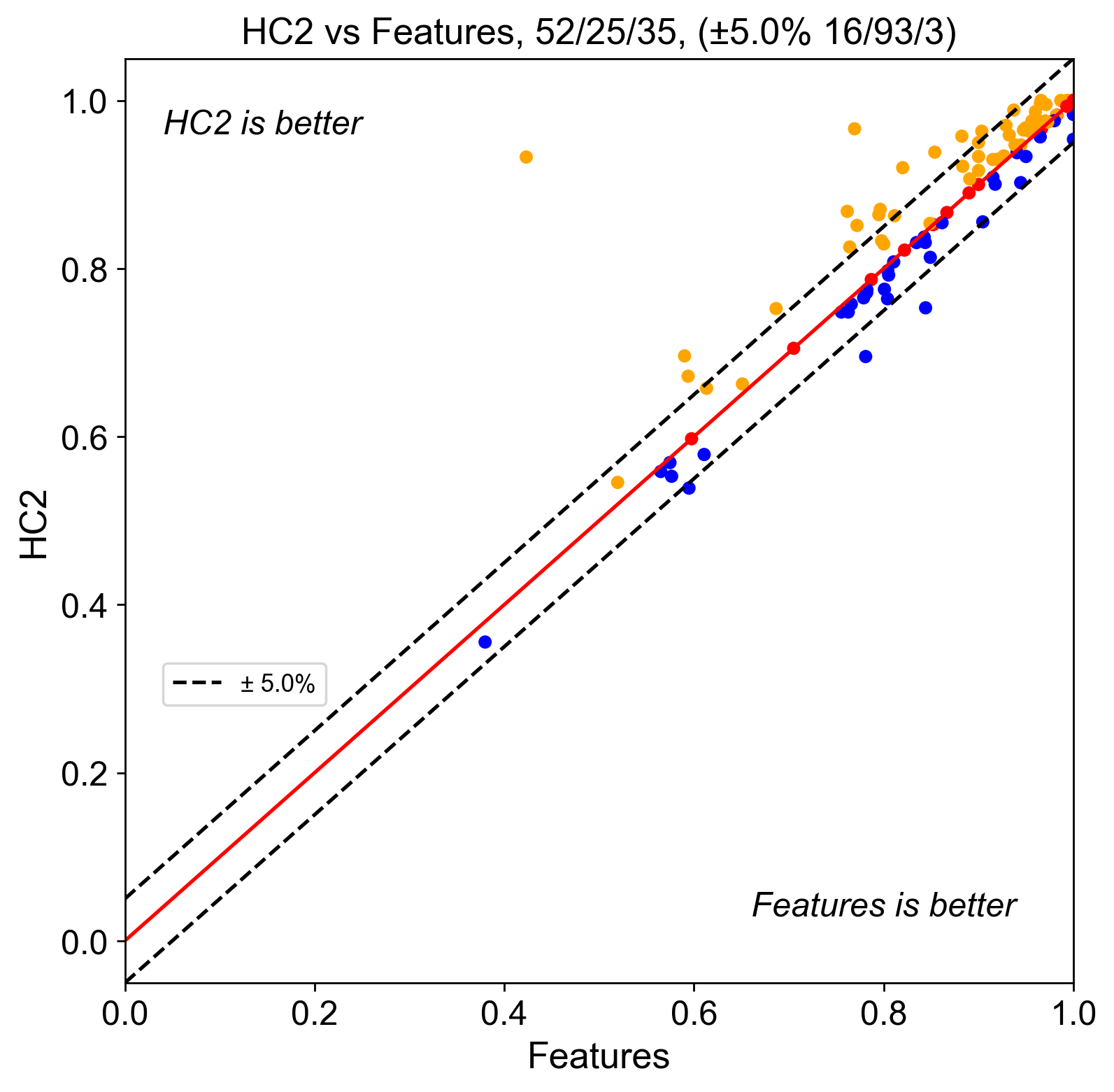}}
    \subfloat{\includegraphics[width=7cm]{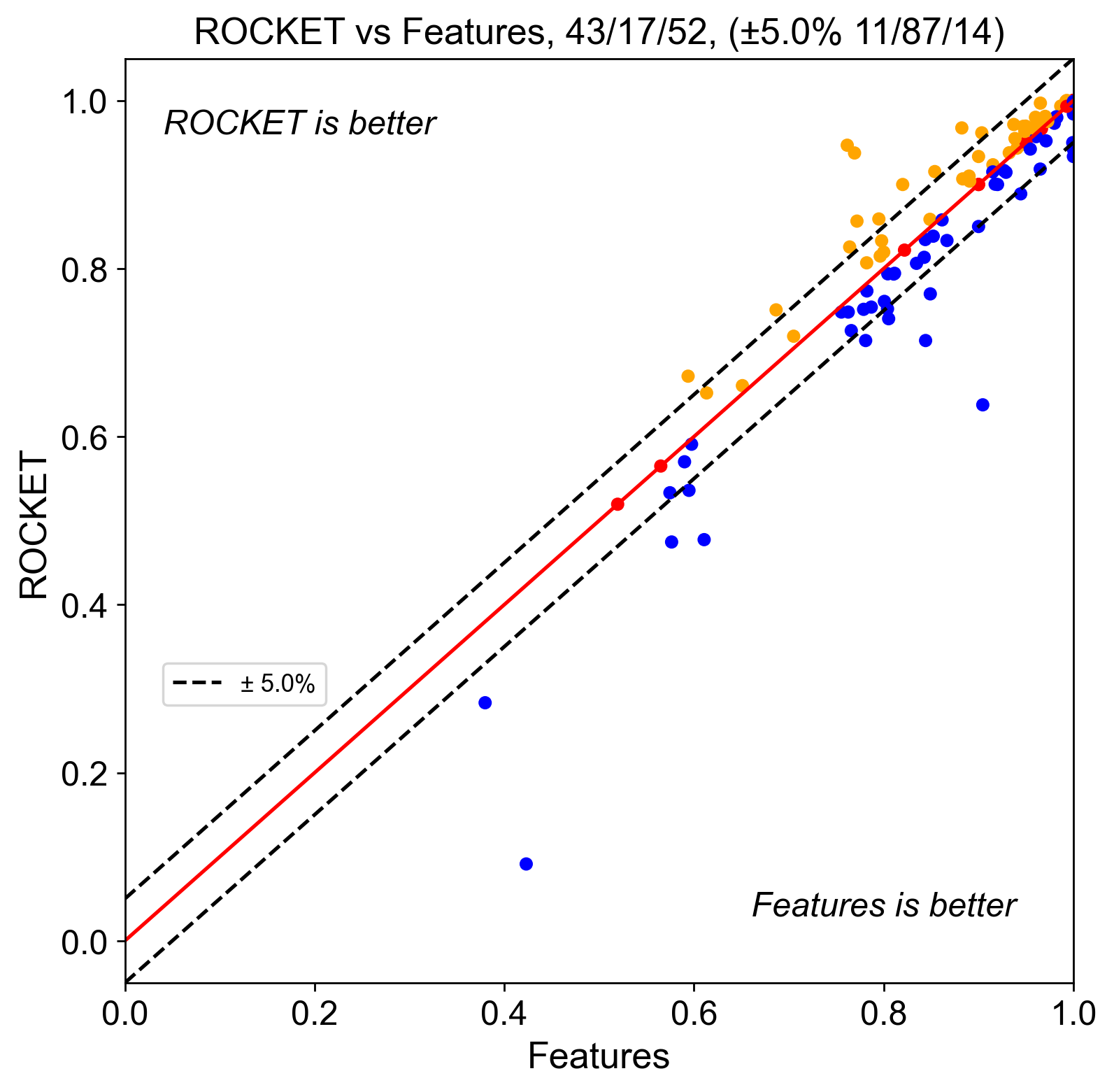}} \\
    \subfloat{\includegraphics[width=7cm]{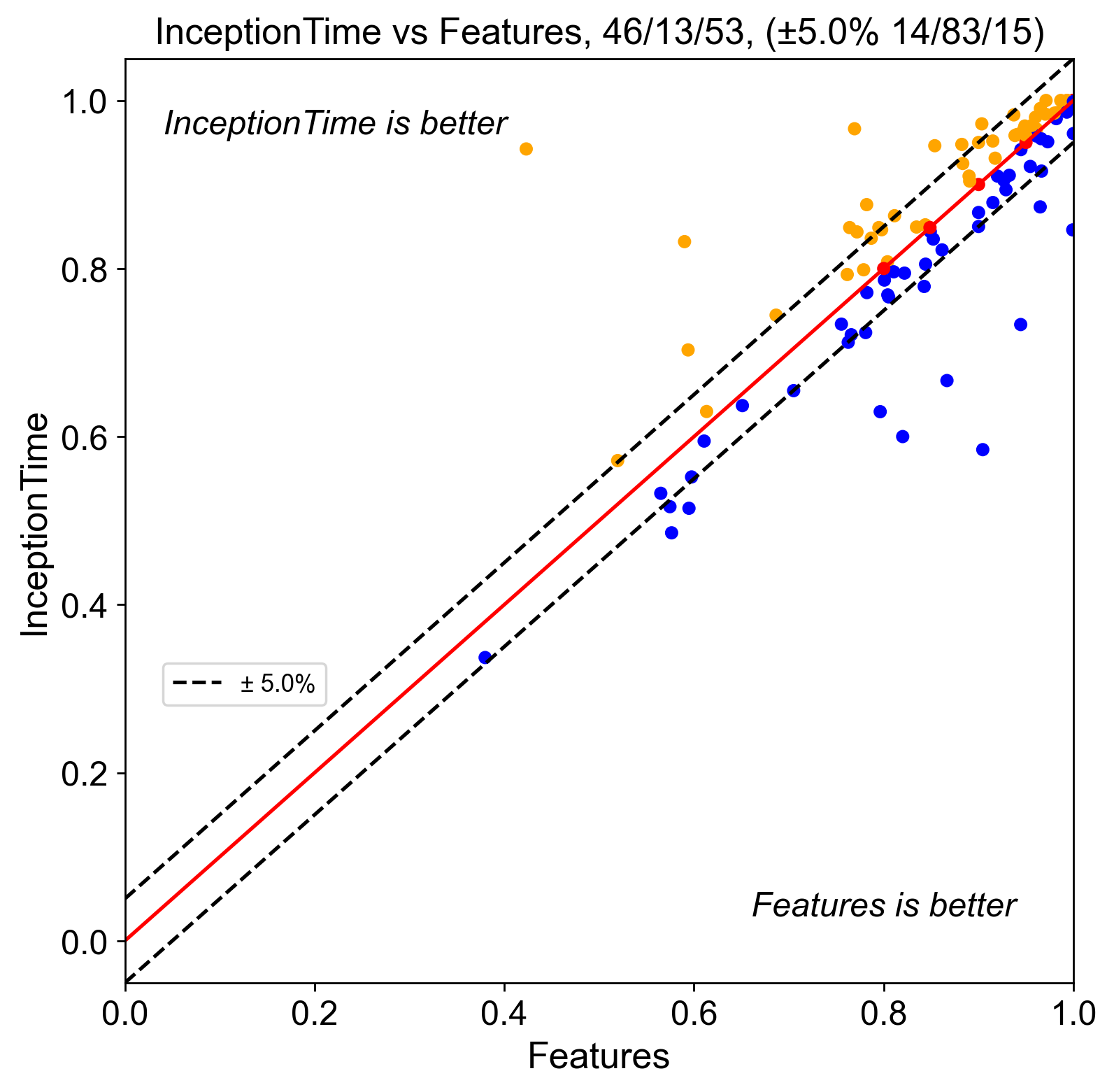}}
    \subfloat{\includegraphics[width=7cm]{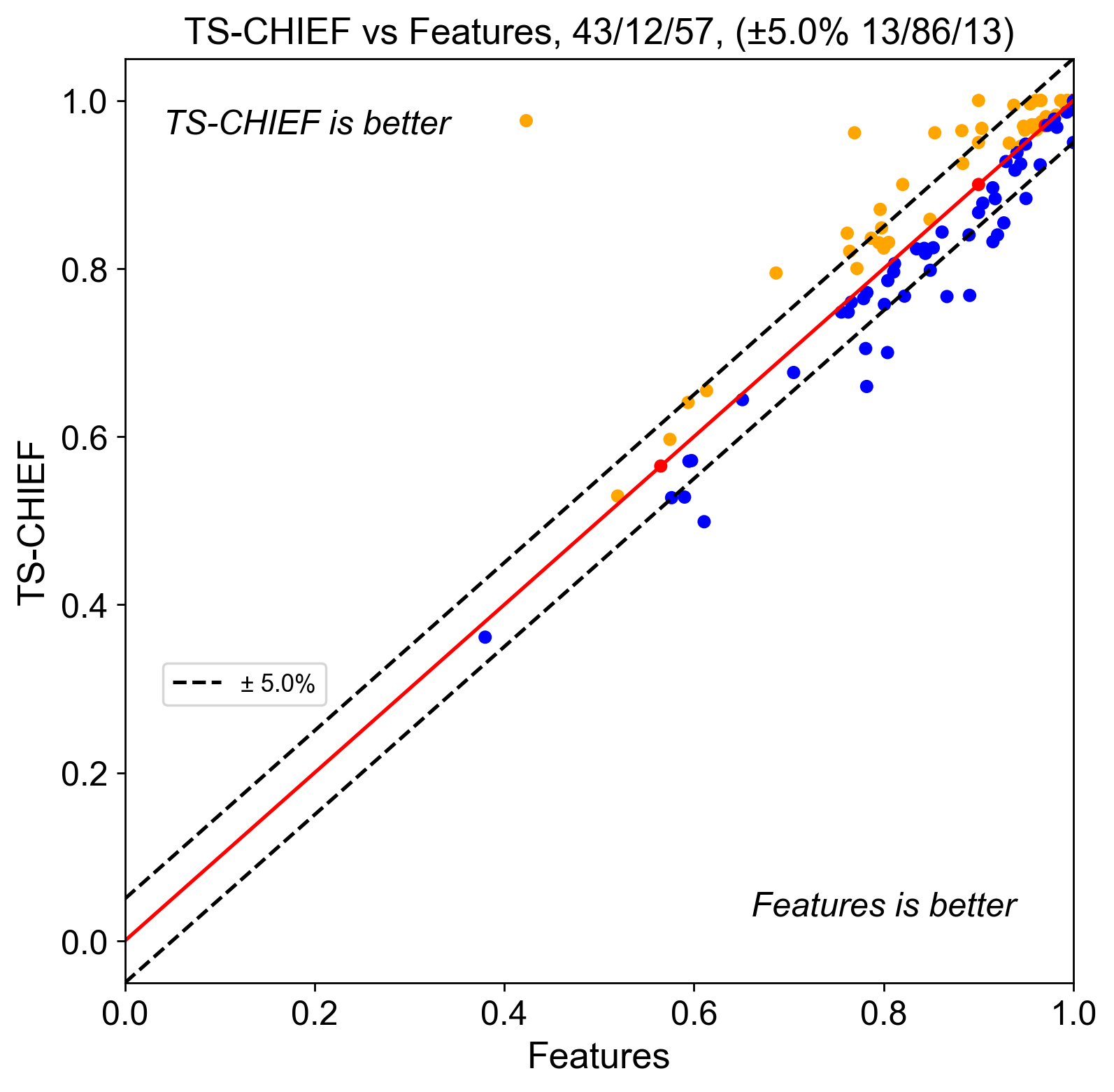}}
    \caption{Scatter plot of accuracy results comparison: our approach ``Features'' vs state of the art TSC algorithms HiveCote2 (HC2), ROCKET, InceptionTime and TS-CHIEF}
    \label{fig:scatter_vs}
\end{figure*}

We are now interested in comparing the obtained performance using our previously defined feature generation libraries with state-of-the-art methods, we can see that our methods are not the worst performers (See Figure \ref{fig:sota}). With the Rotation Forest algorithm (with default parameters) our main approach, noted \textit{Features}, is ranked second among the six others methods, with a the fourth mean accuracy on 112 data sets from the UCR Archive with default train/test splits. Actually, under Nemenyi test, the method is the only one not significantly performing worst than the current most accurate method, namely HIVE-COTE 2.0 (HC2). Figure \ref{fig:scatter_vs} gives additional results by comparing \textit{Features} to four of the best TSC using 1vs1 scatter plots with standard $\pm 5\%$ to highlight notable difference of performance.

As the selected approach contains in itself some SOTA method with MiniROCKET features, we are also providing results when optimizing combining strategy without it. The \textit{Features\_noRocket} approach is, under Holm's correction, as InceptionTime, not significantly worst than HC2: it remains competitive, only loosing $0.2\%$ of mean accuracy comparing to its MiniROCKET version. Replacing MiniROCKET features comes also at the price of around $12$ hours more in total training time from features extraction step.

Concerning scalability, Table \ref{tab:time} demonstrates that the optimal combination of features take a relatively long time to train on the default splits, indeed, it includes the worst scalable methods, i.e. the non-python ones such as \textit{hctsa} and \textit{feasts}. When using only the python available libraries (\textit{Features\_python}), we are roughly dividing the total time by 2, it comes at the cost of exactly $0.1\%$ of average accuracy, not loosing any rank compared with the SOTA classifiers. This approach though, becomes significantly worst than HC2.
Moreover, it is worth-noting that these train time values do not include features extraction step on the test set, which would actually considerably increase the total time (see Table \ref{tab:time_fts}). That being said, even by considering complete feature extraction on both train and test set, our approaches remain substantially faster than HC2 for similar predictive performance on the default split, according to both Nemenyi test and Holm's corrected Wilcoxon signed-rank test.

\begin{table}[!htb]
    \centering
    \begin{tabular}{lll}
        \hline
          \textbf{\small Algorithm} & \textbf{\small Total (h)} & \textbf{\small Average (min)}  \\
        \hline
         ROCKET & 2.85 & 1.53 \\
         \textcolor{red}{Features\_python} & \textcolor{red}{25.15} & \textcolor{red}{13.47} \\
         Arsenal & 27.91 & 14.95 \\
         %DrCIF & 45.40 & 24.32 \\
         \textcolor{red}{Features} & \textcolor{red}{46.16} & \textcolor{red}{24.73} \\
         \textcolor{red}{Features\_noROCKET} & \textcolor{red}{47.62} & \textcolor{red}{25.51} \\
         %TDE & 75.41 & 40.40 \\
         InceptionTime & 86.58 & 46.38 \\
         %STC & 115.88 & 62.08 \\
         HC2 & 340.21 & 182.26 \\
         HC1 & 427.18 & 228.84 \\
         TS-CHIEF & 1016.87 & 544.75 \\
        \hline     
    \end{tabular}
    \caption{\footnotesize Run time to train on a single split on the 112 UCR datasets. For all competitors algorithms, the value for each dataset is the median taken over 30 different resamples. For our approaches, written in red, the value is the one for the default train/test split.}
    \label{tab:time}
\end{table}

\begin{table}[!htb]
    \centering
    \begin{tabular}{llll}
        \hline
          \textbf{\small Algorithm} & \textbf{\small Extraction} & \textbf{\small Extraction} & \textbf{\small Classifier} \\
        & \textbf{\small train (h)} & \textbf{\small test (h)} & \textbf{\small Fit (h)} \\
        \hline
         Features\_python & 15.82 & 37 & 9.33 \\
         Features &  34.05 & 79.58 & 12.11  \\
         Features\_noROCKET & 38.74 & 90.1 & 8.88 \\
        \hline     
    \end{tabular}
    \caption{\footnotesize Run time details of the proposed approaches. Training time is the sum of first and third column.}
    \label{tab:time_fts}
\end{table} 

%% file: conclusion_v2.tex
\section{Conclusion}
\label{conclusion}
In this paper, we aim at exploring the potential predictive power of feature construction (FC) for time series classification (TSC). To this end, we have designed simple processes to branch existing feature engineering tools with standard classifiers. We have led extensive experiments resulting in the comparison of 11 feature engineering tools branched with 9 classifiers over 112 TSC problems -- totalling more than 10000 learning experiments. The analysis of the experimental results indicates that \textit{(i)} using feature engineering leads to better predictive performance for a given standard classifier, \textit{(ii)} combining several unsupervised feature engineering tools with the rotation forest classifier is comparable with the best TSC performers in terms of predictive performance while demanding reasonable computing resources.
The predictive and efficiency results of the suggested approaches also indicate that feature construction fot TSC is a valuable option to pursue. Indeed, it might be naturally extended to multivariate TSC problems -- extracting the same feature set over each dimension or, e.g., randomly sampling some subsets of features in the case of high dimensional data sets. Also, another perspective would be to filter redundant or correlated features to speed up the learning phase.
% This study shows that the usage of general TS designed features seems to be relevant for most of the problems proposed in the UCR benchmark. Thus, one can conclude that approaching SOTA performances only using some simple, easily accessible TS designed tools is possible and that, considering the interpretability and the potential scalability of these kind of methods, those should probably be, at least considered, further in the TSC literature. Moreover those kind of features-based methods can be naturally extended to multivariate TS, extracting the same feature set over each dimension or randomly sampling some subsets of features in the case of high dimensional datasets for example. There is plenty of space in order to optimized both sparsity and computation time of the proposed approaches, we think this could represents some leads for future works.

% \textcolor{magenta}{il faudra faire un petit check des refs + retirer les arxiv pour celles qui sont publiées depuis (mettre les bonnes infos)}